\documentclass[conference]{IEEEtran}

\usepackage{wacv}
\usepackage{times}
\usepackage{epsfig}
\usepackage{graphicx}
\usepackage{amsmath}
\usepackage{amssymb}

\usepackage{algorithm}
\usepackage[noend]{algpseudocode}
\usepackage{subfigure}
\usepackage{multirow}
\usepackage{rotating}
\usepackage{array}

\newcolumntype{L}[1]{>{\raggedright\let\newline\\\arraybackslash\hspace{0pt}}m{#1}}
\newcolumntype{C}[1]{>{\centering\let\newline\\\arraybackslash\hspace{0pt}}m{#1}}
\newcolumntype{R}[1]{>{\raggedleft\let\newline\\\arraybackslash\hspace{0pt}}m{#1}}
\wacvfinalcopy % *** Uncomment this line for the final submission

\def\wacvPaperID{150} % *** Enter the wacv Paper ID here

\ifwacvfinal\fi
\begin{document}

%%%%%%%%% TITLE
\title{Generating Discriminative Object Proposals via Submodular Ranking}

\author{\IEEEauthorblockN{Yangmuzi Zhang\IEEEauthorrefmark{1},
Zhuolin Jiang\IEEEauthorrefmark{2},
Xi Chen\IEEEauthorrefmark{1}, and
Larry S. Davis\IEEEauthorrefmark{1}}
\IEEEauthorblockA{\IEEEauthorrefmark{1}University of Maryland at College Park, MD} \IEEEauthorblockA{\IEEEauthorrefmark{2}Raytheon BBN Technologies, USA}
Email: ymzhang@umiacs.umd.edu}

\maketitle
\ifwacvfinal\thispagestyle{empty}\fi

%%%%%%%%% ABSTRACT
\begin{abstract}
	A multi-scale greedy-based object proposal generation approach is presented. Based on the multi-scale nature of objects in images, our approach is built on top of a hierarchical segmentation. We first identify the representative and diverse exemplar clusters within each scale by using a diversity ranking algorithm. Object proposals are obtained by selecting a subset from the multi-scale segment pool via maximizing a submodular objective function, which consists of a weighted coverage term, a single-scale diversity term and a multi-scale reward term. The weighted coverage term forces the selected set of object proposals to be representative and compact; the single-scale diversity term encourages choosing segments from different exemplar clusters so that they will cover as many object patterns as possible; the multi-scale reward term encourages the selected proposals to be discriminative and selected from multiple layers generated by the hierarchical image segmentation. The experimental results on the Berkeley Segmentation Dataset and PASCAL VOC2012 segmentation dataset demonstrate the accuracy and efficiency of our object proposal model. Additionally, we validate our object proposals in simultaneous segmentation and detection and outperform the state-of-art performance.
\end{abstract}

%%%%%%%%% BODY TEXT
\section{Introduction}
Object recognition has long been a core problem in computer vision. Recent developments in object recognition provide two effective solutions: 1) sliding-window-based object detection and localization~\cite{Viola04, Dalal05, Felzenszwalb10}, 2) segmentation-based approaches~\cite{Carreira12, Uijlings13, Endres14, Arbelaez14}. The sliding window approach incurs high computational cost as it analyses windows over a very large set of locations and scales. Segmentation-based methods lead to fewer regions to consider and to better spatial support for objects of interest with richer shape and contextual information; but the problem of segmenting an image to identify regions with high object spatial support is a challenge.

To improve object spatial support and speed up object localization for object recognition, generating high-quality category-independent object proposals as the input for object recognition system has drawn attention recently~\cite{Endres14, Uijlings13, Cheng14, Arbelaez14}. Motivated by findings from cognitive psychology and neurobiology~\cite{Teuber55, Wolfe04, Desimone95, Koch85} that the human vision system has the amazing ability to localize objects before recognizing them, a limited number of high-quality and category-independent object proposals can be generated in advance and used as inputs for many computer vision tasks. This approach has played a dominant role in semantic segmentation~\cite{Arbelaez12, Carreira12ECCV} and leads to competitive performance on detection~\cite{Fidler13}. There are two main categories of object proposal generation methods depending on the shape of proposals: bounding-box-based  proposals~\cite{Zitnick14, Cheng14, Uijlings13} and segment-based proposals~\cite{Arbelaez14, Endres14, Singh14}. %The latter provides more accurate shape and location details of objects of interest.

\begin{figure*}
\begin{center}
%\hspace{-0.3cm}
\subfigure[Input]
{
\label{fig:ob:a}
\includegraphics[width=0.22\linewidth]{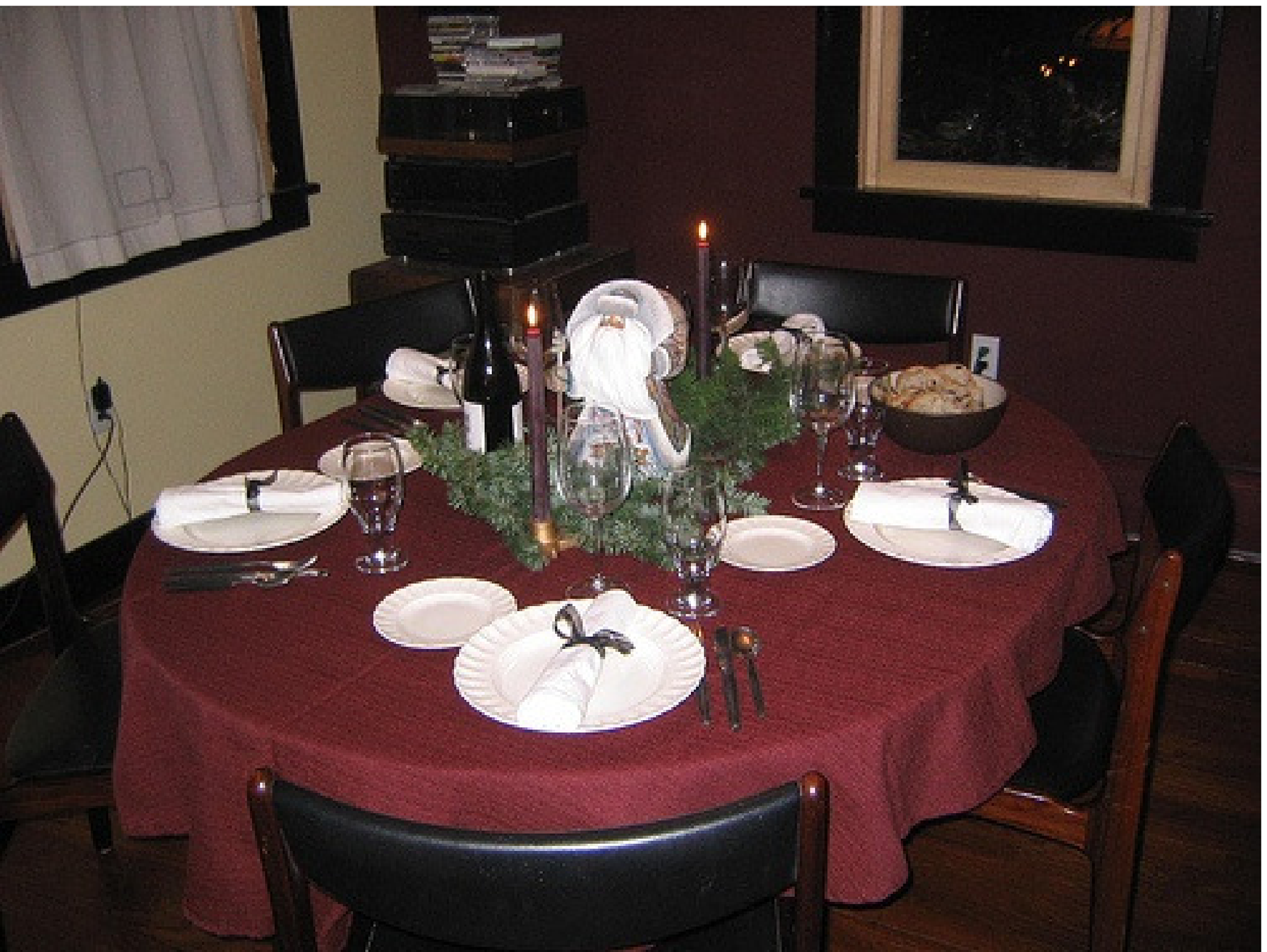}
}
%\hspace{-0.3cm}
\subfigure[Coarse layer sample]
{
\label{fig:ob:b}
\includegraphics[width=0.22\linewidth]{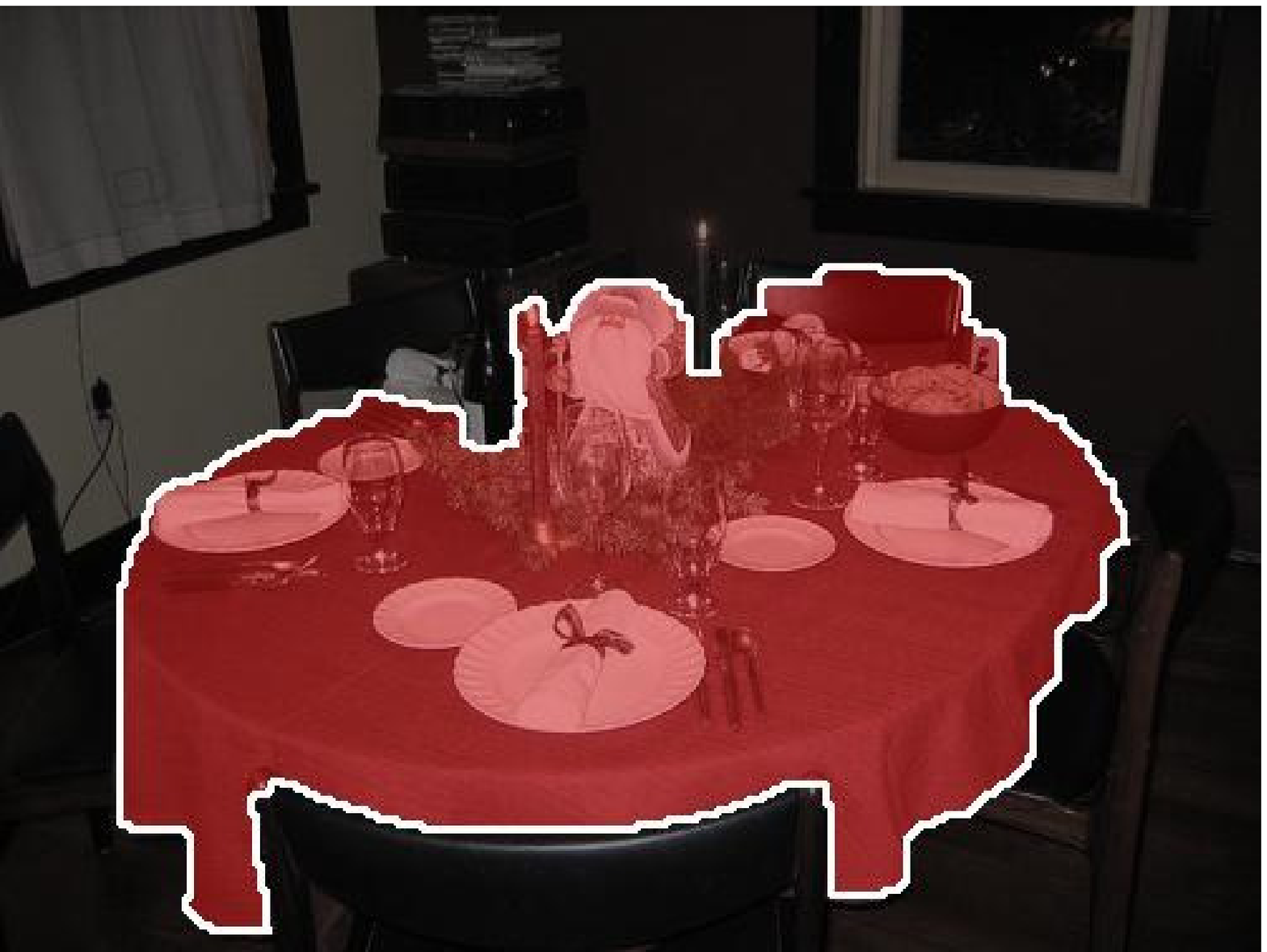}
}
%\hspace{-0.3cm}
\subfigure[Middle layer samples]
{
\label{fig:ob:c}
\includegraphics[width=0.22\linewidth]{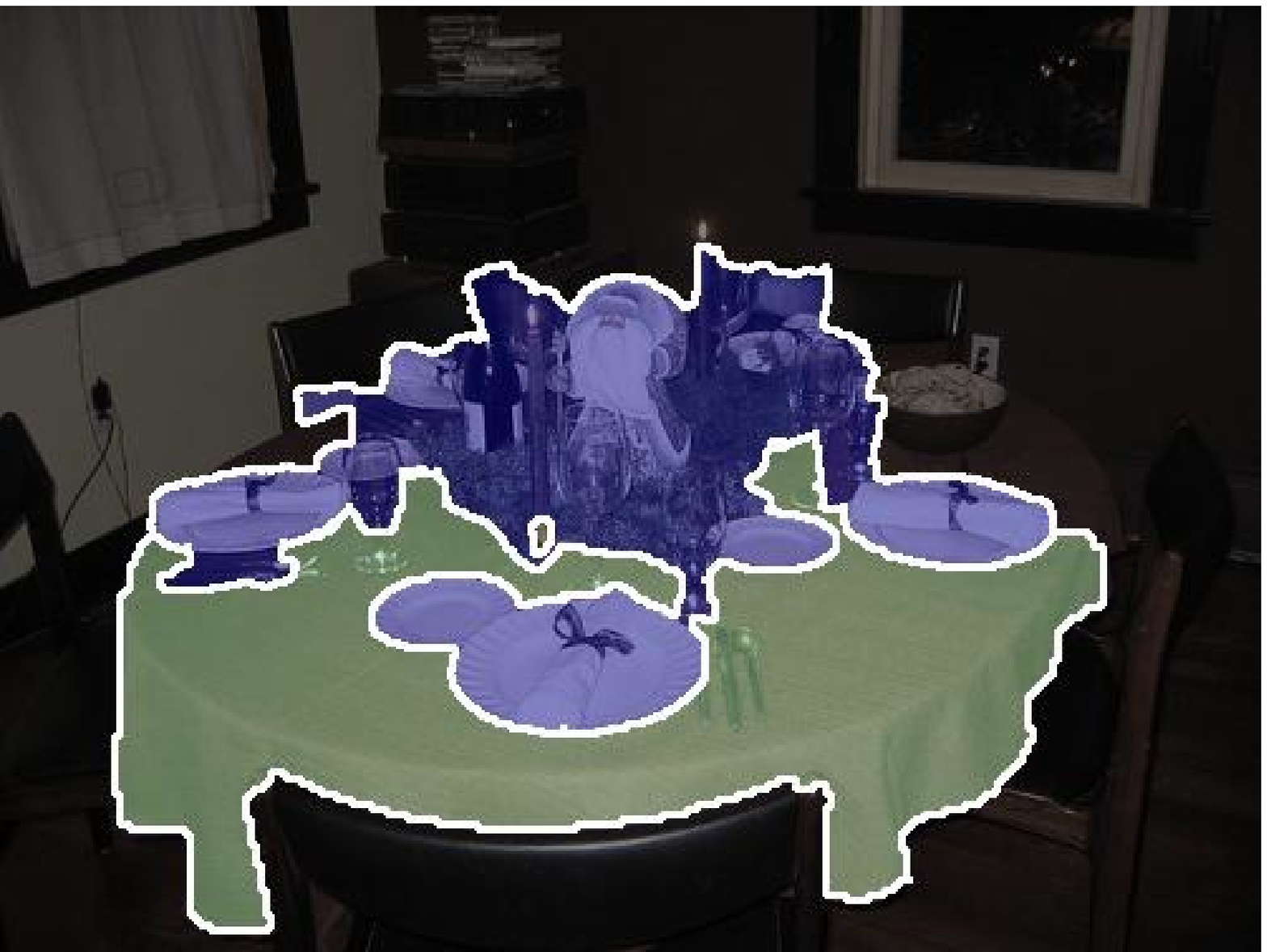}
}
%\hspace{-0.3cm}
\subfigure[Fine layer samples]
{
\label{fig:ob:d}
\includegraphics[width=0.22\linewidth]{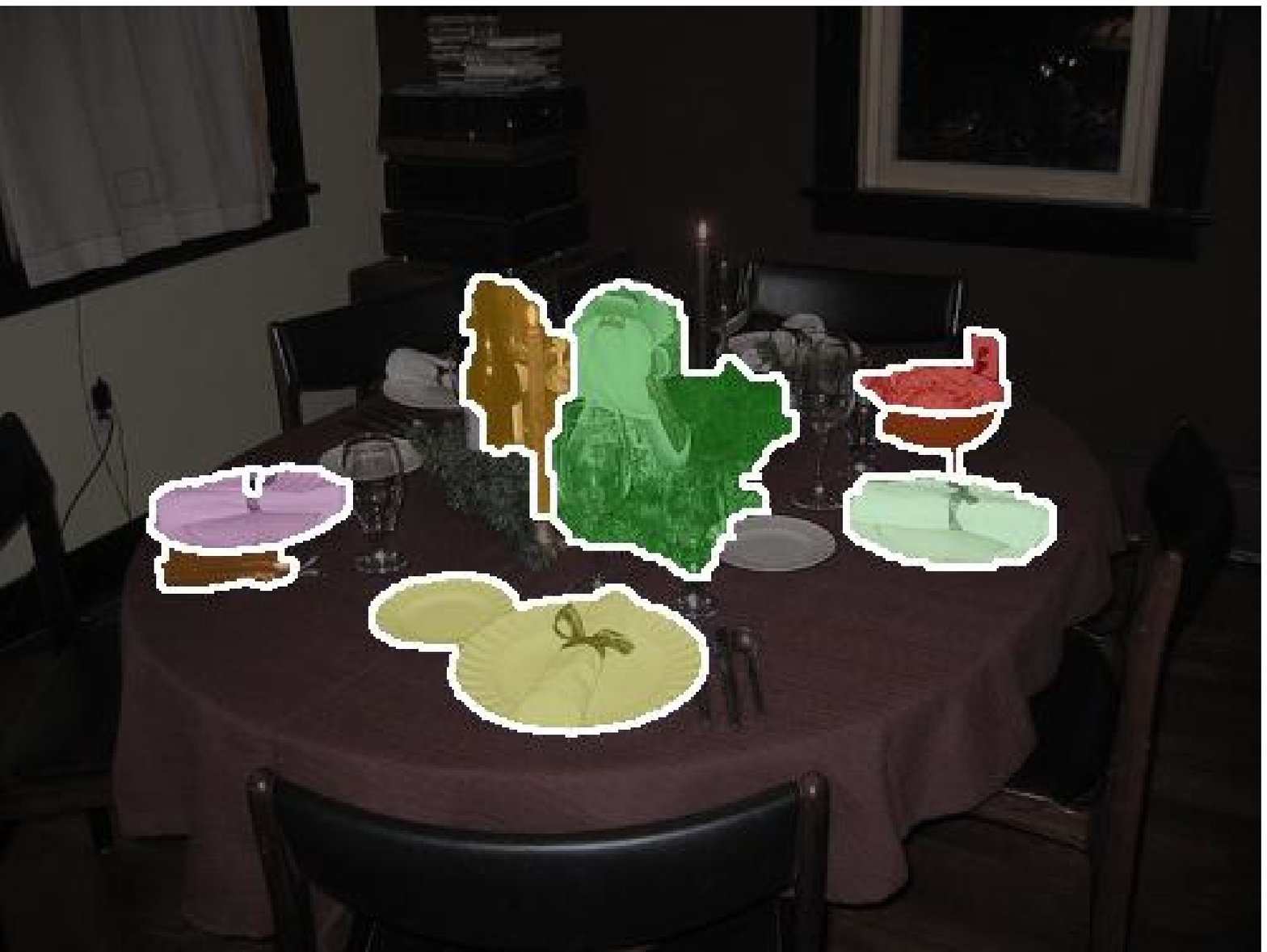}
}\\
%\subfigure[]
%{
%\label{fig:ob:e}
%\includegraphics[width=0.18\linewidth]{fig/img6241/2007_006241_gt.jpg}
%}
\subfigure[CPMC~\cite{Carreira12}]
{
\label{fig:ob:e}
\includegraphics[width=0.22\linewidth]{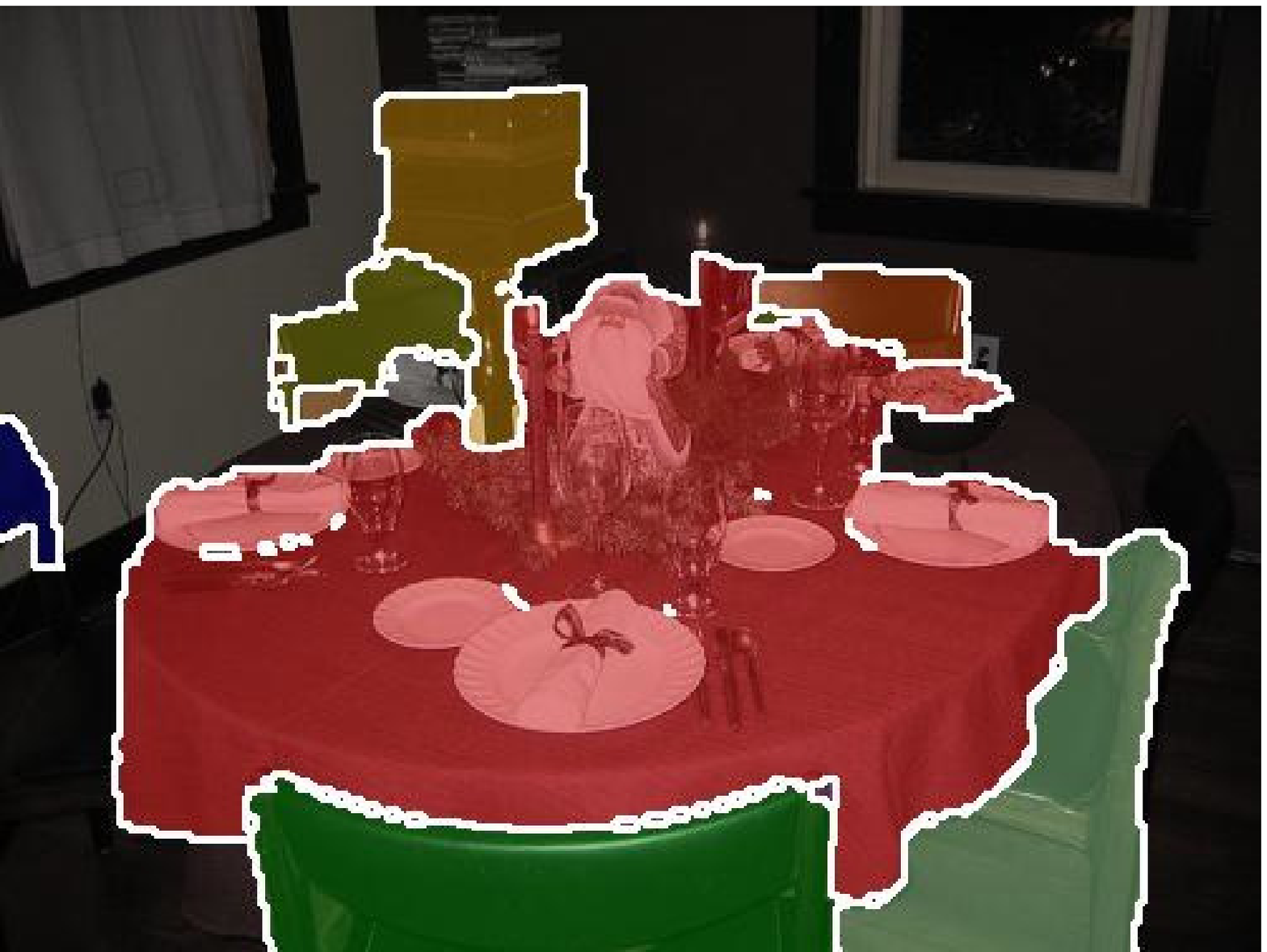}
}
\subfigure[Categ. Indep.~\cite{Endres14}]
{
\label{fig:ob:f}
\includegraphics[width=0.22\linewidth]{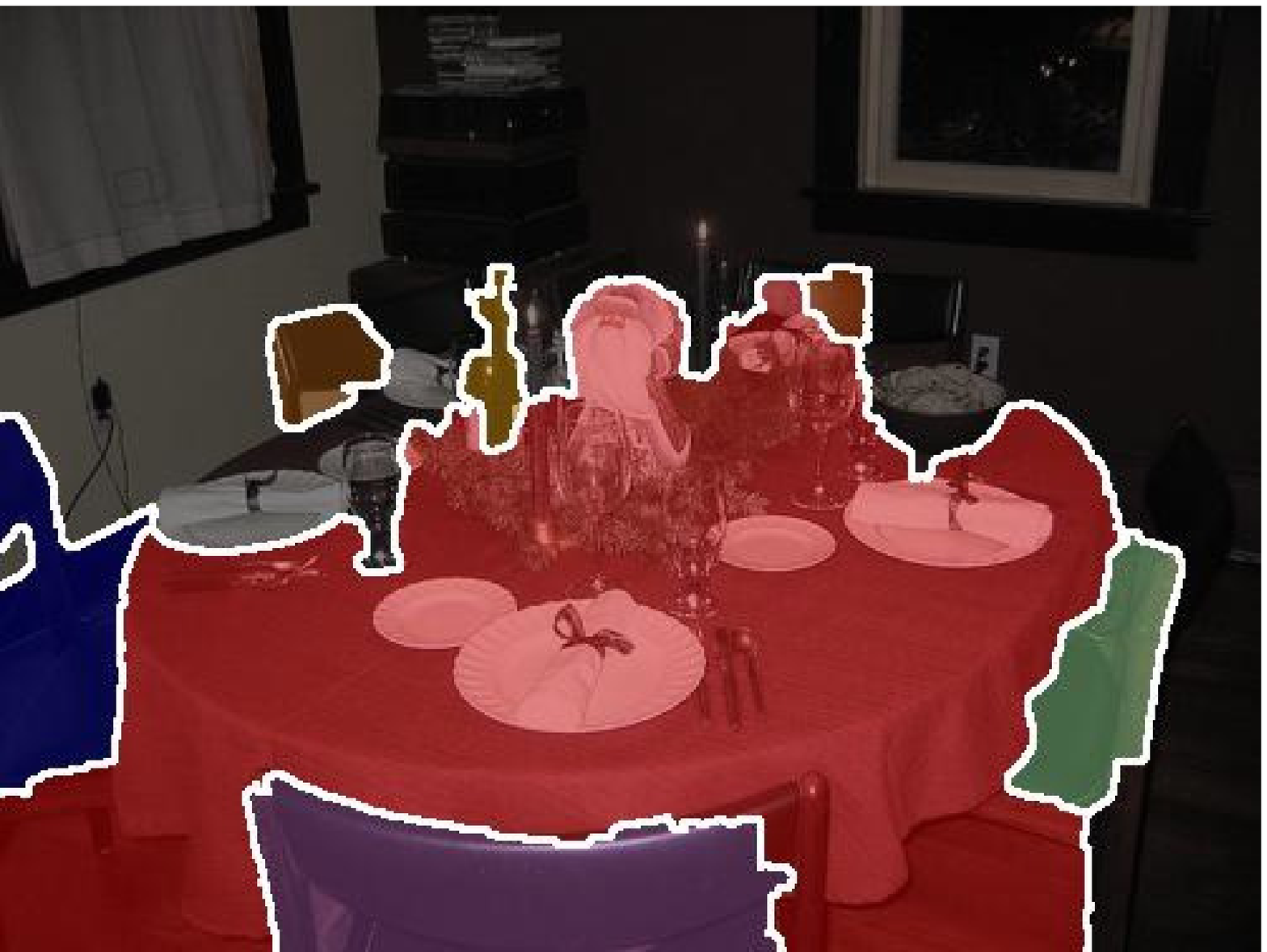}
}
\subfigure[MCG~\cite{Arbelaez14}]
{
\label{fig:ob:g}
\includegraphics[width=0.22\linewidth]{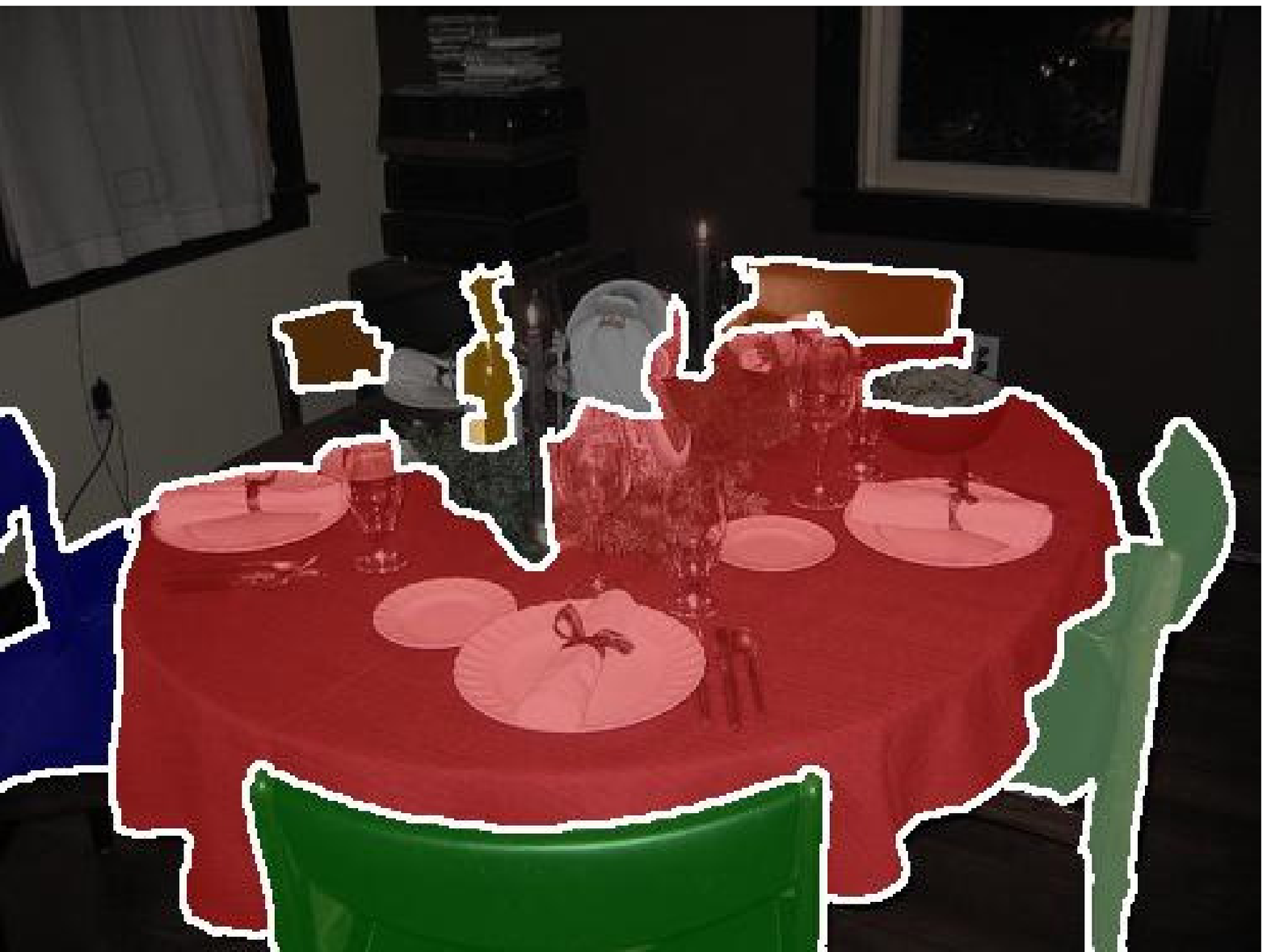}
}
\subfigure[Our method]
{
\label{fig:ob:h}
\includegraphics[width=0.22\linewidth]{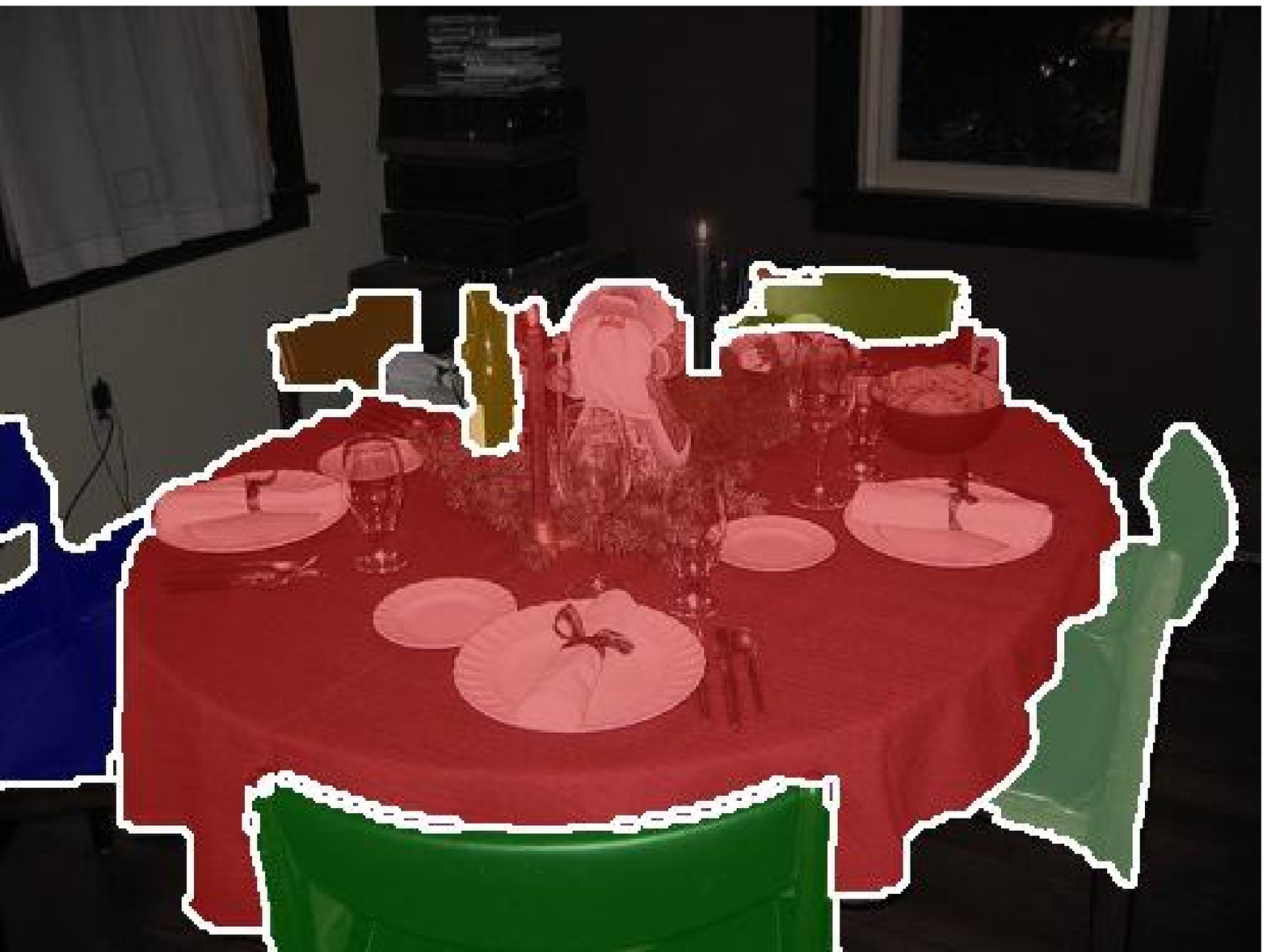}
}
\end{center}
%\vspace{-0.1cm}
\caption{Objects in an image are naturally hierarchical. (a) is an original image from Pascal VOC2012; (b) - (d) show segments around the table from different scales using method~\cite{Carreira12}; (e) shows the best seven object proposals generated from CPMC~\cite{Carreira12}; (f) are proposals from Categ. Indep.~\cite{Endres14}; (g) are proposals from MCG~\cite{Arbelaez14}; (h) are proposals from our method.}
\label{fig:ob}
%\vspace{-0.1cm}
\end{figure*}

Objects in an image are intrinsically hierarchical and of different scales. Consider the table in Figure~\ref{fig:ob:a} for example. The objects on the table can be regarded as a part of the table (Figure~\ref{fig:ob:b}), and at the same time, they constitute a group of objects on the table (Figure~\ref{fig:ob:c}). More specifically, these objects include plates, forks, the Santa Claus, and a bottle (Figure~\ref{fig:ob:d}). Therefore, multi-scale segmentation is essential to localize and segment different objects. There have been a few attempts~\cite{Carreira12, Endres14, Arbelaez14} to combine multiple scale information in the object proposal generation process, but very few papers have studied the importance of proposal selection given segments from hierarchical image segmentations. Figure~\ref{fig:ob:e}\ref{fig:ob:f}\ref{fig:ob:g} show the generated proposals from three state-of-art algorithms~\cite{Carreira12, Endres14, Arbelaez14}. However, they do not cover all the objects in the image well.% As shown in the second row of Figure~\ref{fig:ob}, compared to the existing methods (Figure~\ref{fig:ob:e} - \ref{fig:ob:g}), our method (Figure~\ref{fig:ob:h}) can properly select representative, diverse and discriminative object proposals from different layers ( for example, the bottle from finer layer and the table from raw layer).
 
%\begin{figure}
%\begin{center}
%\subfigure[]
%{
%\label{fig:ob:a}
%\includegraphics[width=0.55\linewidth]{fig/2007_006241.jpg}
%}
%\subfigure[]
%{
%\label{fig:ob:b}
%\includegraphics[width=0.31\linewidth]{fig/2007_008110.jpg}
%}\\
%\subfigure[]
%{
%\label{fig:ob:c}
%\includegraphics[width=0.52\linewidth]{fig/2011_000548.jpg}
%}
%\subfigure[]
%{
%\label{fig:ob:d}
%\includegraphics[width=0.38\linewidth]{fig/2011_001060.jpg}
%}
%\end{center}
%\caption{Objects in an image are naturally hierarchical.}
%\label{fig:ob}
%\vspace{-0.4cm}
%\end{figure}

We present a greedy approach to efficiently extract high-quality object proposals from an image via maximizing a submodular objective function. We first construct diverse exemplar clusters of segments over a range of scales using diversity ranking; then rank and select high-quality object proposals from the multi-scale segment pool generated by hierarchical image segmentation. Our objective function is composed of three terms: a weighted coverage term, a single-scale diversity term and a multi-scale reward term. The first term encourages the selected set to be compact and well represent all segments in an image. The second term enforces the selected segments (object proposals) to be diverse and cover as many different objects as possible. The third term encourages the selected proposals to correspond to objects with high confidence and selected from different scales. The algorithm takes object scale information into account and avoids selecting segments from the same layer repeatedly. Compared to existing segment-based methods, our method (Figure~\ref{fig:ob:h}) can select representative, diverse and discriminative object proposals from different layers (for example, the bottle from fine layer and the table from coarse layer). Our main contributions are as follows:

\begin{itemize}
\item The generation of object proposals is solved by maximizing a submodular objective function.  An efficient greedy-based optimization algorithm with guaranteed performance is presented based on the submodularity property. 
%We present an efficient greedy algorithm to solve the objective function based on its submodularity property. This algorithm provides a performance-guaranteed solution. 
%This provides us with a computationally efficient algorithm. %This introduces a new perspective of using submodularity to produce object proposals.
\item We naturally integrate multi-scale and object discriminativeness information into the objective function. The generated proposals are representative, diverse and discriminative.%and it can be efficiently solved due to its submodularity property.
%\item We present an efficient greedy algorithm to solve the objective function based on its submodularity property. This algorithm can provide a performance-guaranteed solution. 
\item Our approach achieves state-of-the-art performance on two popular datassets, and our generated object proposals, when integrated into simultaneous segmentation and detection, achieves state of the art results.
\end{itemize}

%We have evaluated our method on the BSDS dataset and PASCAL VOC2012 segmentation dataset. The experimental results show that our method achieves state-of-art object-level accuracy while computationally more efficient. We further test our object proposals with simultaneous detection and segmentation task~\cite{Hariharan14} and achieve higher AP score than object proposals generated from other existing methods.

%-------------------------------------------------------------------------
\section{Related work}
The goal of object proposal algorithms is to generate a small number of high-quality category-independent proposals such that each object in an image is well captured by at least one proposal~\cite{Alexe12, Endres14}.
%Compared to the sliding window paradigm which requires a sophisticated classifier for each category, it reduces both computational and model-learning cost dramatically by decreasing the candidate pool to which class models need to be applied. \rk{in object proposal, we learn classifier for object and non-object instead of for each object categories.} One group of methods produces bounding-box-based object proposals. 
Existing object proposal approaches can be roughly divided into bounding-box and segment based approaches. \cite{Zitnick14} generated bounding boxes by utilizing edge and contour clues. In~\cite{Uijlings13}, a data-driven grouping strategy which combines segmentation and exhaustive search is presented to produce bounding-box-based proposals. \cite{Cheng14} proposed the binarized normed gradients (BING) feature to efficiently produce object boxes. Instead of generating bounding-box-based proposals, our work focuses on extracting segment-based proposals which aims to cover all the objects in an image and can provide more accurate shape and location information. Some algorithms have been reported to generate segment-based object proposals. \cite{Carreira12} segmented objects by solving a series of constrained parametric min-cut (CPMC) problems. \cite{Humayun14} reused inference in graph cuts to solve the parametric min-cut problems much more efficiently. \cite{Endres14} performed graph cuts and ranked proposals using structured learning. In~\cite{Arbelaez14}, a hierarchical segmenter is used to combine multi-scale information, and a grouping strategy is presented to extract object candidates. Different from their work, we design an efficient greedy-based ranking method to leverage multi-scale information in the process of selecting object proposals from a large hierarchical segment pool. %(To the best of our knowledge, we are the first to explore multi-scale object proposal generation using a submodular ranking model.) \rk{Shall I add the last sentence?}

Object proposals have been used in many computer vision tasks, such as segmentation~\cite{Arbelaez12, Carreira12}, object detection~\cite{Fidler13} and large-scale classification~\cite{Uijlings13}. Semantic segmentation and object detection have been shown to support each other mutually in a wide variety of algorithms. \cite{malisiewicz-bmvc07} showed that better quality segmentation can improve object recognition performance.~\cite{Fidler13, Chen11, Hariharan14} used hierarchical segmentations and combined several top-down cues for object detection. The more demanding task of simultaneous detection and segmentation (SDS) is investigated in~\cite{Hariharan14} which detects and labels the segments at the same time. We use this same detection and segmentation framework but with our object proposal generation method to demonstrate the effectiveness of proposals generated by our approach.

%{\bf Convolutional Neural Network} Convolutional Neural Network (CNN) was popularly used for object recognition in applications like digit recognition~\cite{lecun1989backpropagation} using backpropagation. The substantial improvement of state-of-the-art results on ImageNet Large Scale Visual Recognition Challenge (ILSVRC)~\cite{deng2012imagenet} using deep CNN structure in~\cite{krizhevsky2012imagenet} draws great attentions to the field in both academia and industry. In recent years, using CNN features extracted and fine-tuned on regions generated by object proposal algorithms has been a popular pipeline and has achieved good results on object recognition benchmarks. ~\cite{girshick14CVPR} proposed the R-CNN framework to generalize the~\cite{krizhevsky2012imagenet} model to object detection in PASCAL VOC Challenge by classifying region proposals generated by~\cite{Sande11} to achieve state-of-the-art object detection results. ~\cite{Hariharan14} recently extends the R-CNN framework to simultaneous object segmentation and detection. It learns and extracts features using object proposals from~\cite{Arbelaez14} and shows promising results on the SDS tasks on PASCAL VOC2012 dataset.

Submodular optimization is a useful optimization tool in machine learning and computer vision problems~\cite{Leskovec07, Liu13, Kim12CVPR, Jiang13, Liu14, Zhu14}. \cite{Leskovec07} demonstrates how submodularity speeds up optimization algorithm in large scale problems. In~\cite{Kim12CVPR}, a diffusion-based framework is proposed to solve cosegmentation problems via submodular optimization. \cite{Jiang13} used the facility location problem to model salient region detection where salient regions are obtained by maximizing a submodular objective function.

\section{Submodular Proposal Extraction}
We first obtain a large pool of segments from different scales using hierarchical image segmentation. Diverse exemplar clusters are then generated via diversity ranking within each layer to discover potential objects in an image. We define a submodular objective function to rank and select a discriminative and compact subset from a large set of segments of different scales, then the selected segments are used as the final object proposals.

\subsection{Preliminaries}
\textbf{Submodularity:} Let $V$ be a finite set, $A\subseteq B \subseteq V$ and $a \in V\setminus B$. A set function $F: 2^v \rightarrow R$ is submodular if $F(A\bigcup a)-F(A)\geqslant F(B\bigcup a)-F(B)$. This is the diminishing return property: adding an element to a smaller set helps more than adding it to a larger set~\cite{Nemhauser78}.

\subsection{Hierarchical Segmentation}
We build our object proposal generation framework on top of hierarchical segmentation. Following~\cite{Carreira12, Humayun14}, we generate segments for an image at different scales by solving multiple constrained parametric min-cut problems with different seeds and unary terms. %One can also generate hierarchical segmentations based on the output boundary algorithm~\cite{Hoiem11,Endres14}.

\subsection{Exemplar Cluster Generation}
\label{sec:pce}
In a coarser layer, an image is segmented into only a few segments. However, the number of segments increases dramatically as we go to finer layers. To reduce the redundancy and maintain segment diversity, we introduce an exemplar cluster generation step to pre-process segments within layers. %If the number of segments of a layer $l$ is greater than a threshold $N_t$ ($N_t=800$), we will employ diversity ranking and agglomerative clustering~\cite{Kim12CVPR} on segments within that layer.

Let $V$ denote the set containing segments from all layers of an image (the multi-scale segment pool), and $V^l$ be the set of segments from layer $l$. Then $V = \bigcup_{l=1}^L V^l$, $L$ is the total number of layers, and $V^l$s are disjoint. For each layer $l$, we obtain a partition of its segments $\{P_1^l, P_2^l, ..., P_t^l\}$ using a diversity ranking algorithm~\cite{Kim12CVPR}. $P_t^l$ is the set of segments assigned to cluster $t$. Each segment belongs to only one cluster, and clusters are disjoint. For each layer $L$, we have $V^l=\cup _{t=1}^{T}P_t^l$, where $T$ is the number of clusters\footnote{For coarser layer, $T$ is the number of initial segments obtained from hierarchical segmentation.}. %Cluster centres from all layers form the object candidate set $J$.

\subsection{Submodular Multi-scale Proposal Generation}
%Usually, objects are of different scales in an image. It is difficult to pull them out simultaneous using one scale. In addition, it will be costly if using segments from all different layers as object proposals. Therefore, 
We present a proposal generation method by selecting a subset $A$ which contains high-quality segments (object proposals) from the set $V$. %As $V$ contains a large amount of object candidates, our goal is to select a small set of segments where each segment serves as a high-quality and category-independent object proposal.

\begin{figure}
\begin{center}
%\hspace{-0.3cm}
%\subfigure[H(A)=5.9]
%{
%\label{fig:wc:a}
%\includegraphics[width=0.3\linewidth]{fig/WC_v2_1.png}
%}
\subfigure[H(A)=7.7]
{
\label{fig:wc:a}
\includegraphics[width=0.47\linewidth]{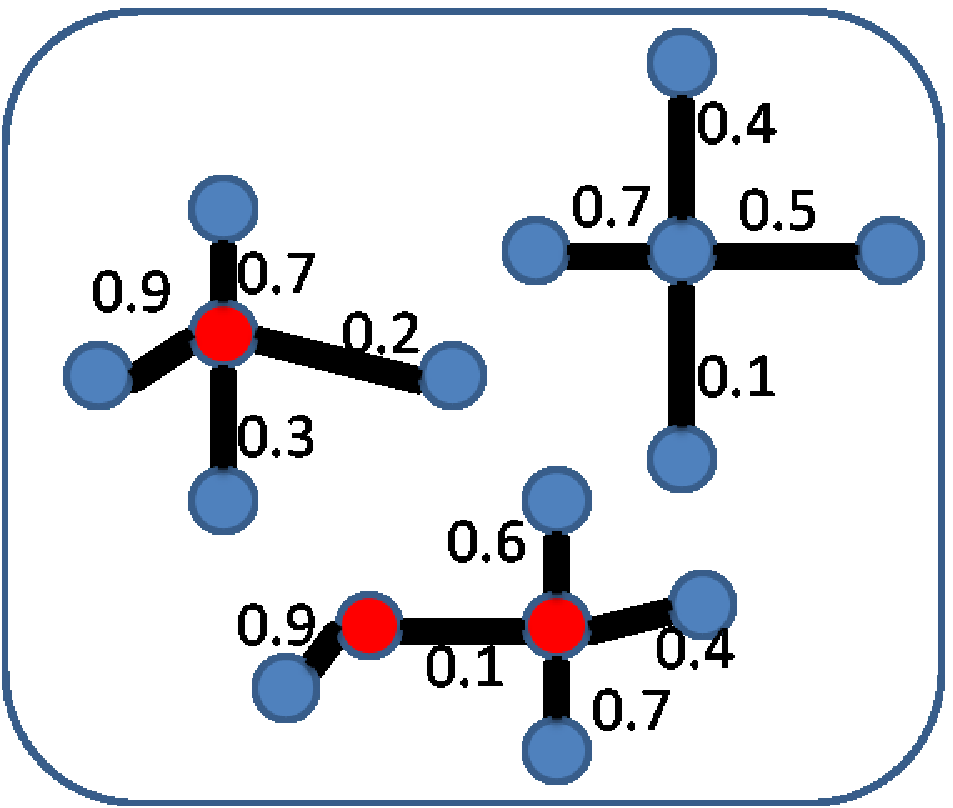}
}
%\hspace{-0.3cm}
\subfigure[H(A)=8.6]
{
\label{fig:wc:b}
\includegraphics[width=0.47\linewidth]{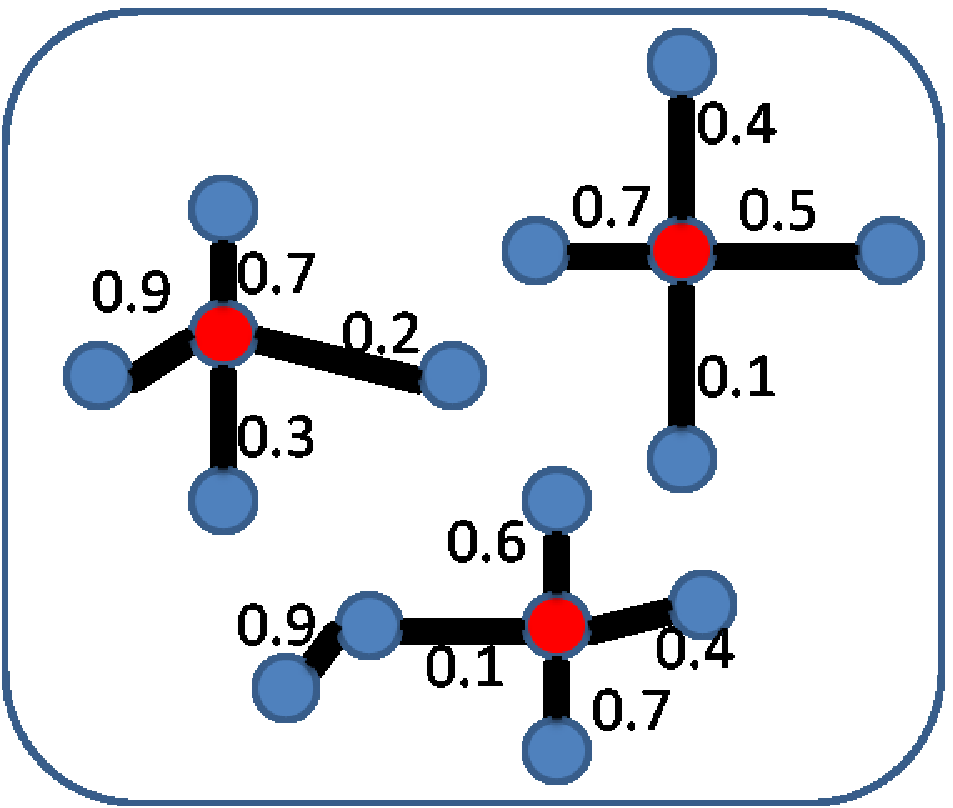}
}
%\hspace{-0.3cm}
%\subfigure[]
%{
%\label{fig:fl:d}
%\includegraphics[width=0.47\linewidth]{fig/fl4.png}
%%}
\end{center}
%\vspace{-0.1cm}
\caption{The weighted coverage term for the representative proposal selection (best viewed in color). The node denotes the segment vertex, and the value next to the edge is the similarity between vertices. The red nodes are selected vertices. To select three nodes among all, by computing the weighted coverage term, we favours selecting a more representative set (three center nodes in (b) will lead to higher $H(A)$ than the less representative one since the two nodes are from one group in (a)). Hence the selected $A$ is representative and compact.}
\label{fig:wc}
%\vspace{-0.1cm}
\end{figure}

Given an image $I$, we construct an undirected graph $G=(V,E)$ for the segment hypotheses in $I$. Each vertex $v\in V$ is an element from the multi-scale segment pool. Each edge $e \in E$ models the pairwise relation between vertices. Two segments are connected if they are overlapping (between layers) or adjoining (within a layer). The weight $w_{ij}$ associated with the edge $e_{ij}$ measures the appearance similarity between vertices $v_i$ and $v_j$. We extract a CNN feature descriptor~\cite{Girshick14CVPR} for
each segment: $X=[x_1,x_2,...,x_{|V|}]$. $w_{ij}$ is defined as the Gaussian similarity between two vertices' feature descriptors.
\begin{equation}
  w_{ij}=\begin{cases}
    exp(-\epsilon d^2(x_i,x_j)), & \text{if $e_{ij}\in E$}.\\
    0, & \text{otherwise}.
  \end{cases}
\end{equation}
As suggested in~\cite{Zelnik04}, we set the normalization factor $\epsilon=1/\sigma_i \sigma_j$ and the local scale $\sigma_i$ is selected by the local statistic of vertex $i$'s neighbourhood. We adopt the simple choice which sets $\sigma_i=d(x_i,x_M)$ where $x_M$ corresponds to the $M$'th closest neighbour of vertex $i$.

\subsubsection{Weighted Coverage Term}
\label{subsubsec:wc}
The selected subset $A$ should be representative of the whole set $V$. The similarity of subset $A$ to the whole set $V$ is maximized with a constraint on the size of $A$. Accordingly, we introduce a weighted coverage term for selecting representative proposals.

%Given an image $I$, we construct a graph $G(V,E)$ based on the patch hypotheses in image $I$. Each vertex $v\in V$ is an element in the multi-scale object candidate pool obtained from the previous step. Each edge $e \in E$ models the pairwise relation between vertices. The weight $w_{ij}$ associated with the edge $e_{ij}$ measures the similarity between vertices $i$ and $j$. We extract CNN feature descriptor for each object candidate: $X=[x_1,x_2,...,x_N]$, where $N$ is the number of object candidates. $w_{ij}$ is defined as Gaussian similarity between two vertices' feature descriptors.
%\begin{equation}
%  w_{ij}=\begin{cases}
%    exp(\epsilon d^2(x_i,x_j)), & \text{if $e_{ij}\in E$}.\\
%    0, & \text{otherwise}.
%  \end{cases}
%\end{equation}
%As suggested in~\cite{Zelnik04}, we set normalization factor $\epsilon=1/\sigma_i \sigma_j$ and the local scale $\sigma_i$ is selected by the local statistic of vertex $i$'s neighbourhood. We adopt the simple choice which sets $\sigma_i=d(x_i,x_M)$ and $x_M$ corresponds to $M$'th neighbour of vertex $i$.

Let $N_A$ denote the number of selected segments. Then the weighed coverage term is formulated as:
\begin{eqnarray}
\label{eqn:facility}
H(A)=\sum_{i\in V}\max_{j \in A}w_{ij} \\
s.t. \quad A \subseteq V, N_A\leqslant K \nonumber
%s.t. \quad A \subseteq J \subseteq V, N_A\leqslant K \nonumber
\end{eqnarray}
where $K$ is the maximum number of segments to be chosen in set $A$. The weighted coverage of each segment $v_i$ is $\max_{j \in A}w_{ij}$. Equation~(\ref{eqn:facility}) measures the representativeness of $A$ to $V$ and favours selecting segments which can cover (or represent) the other unselected segments. Maximizing the weighted coverage term encourages the selected set $A$ to be representative and compact as shown in Figure~\ref{fig:wc}.

%\begin{figure}
%\begin{center}
%   \includegraphics[width=0.9\linewidth]{fig/WS.png}
%\end{center}
%   \caption{Object proposal quality at instance level on PASCAL VOC2012 validation set.}
%\label{fig:WC}
%\vspace{-0.5cm}
%\end{figure}

\subsubsection{Single-Scale Diversity Term}
The weighted coverage term will give rise to a highly representative set $A$; however, segments from each layer (corresponding to each image scale) still possess redundancy. Therefore, we introduce a diversity term to force segments within a layer $l$ to be different. The single-layer diversity term is formulated as follows:
\begin{figure}
\begin{center}
\subfigure[D(A)=0.90]
{
\label{fig:SD:a}
\includegraphics[width=0.47\linewidth]{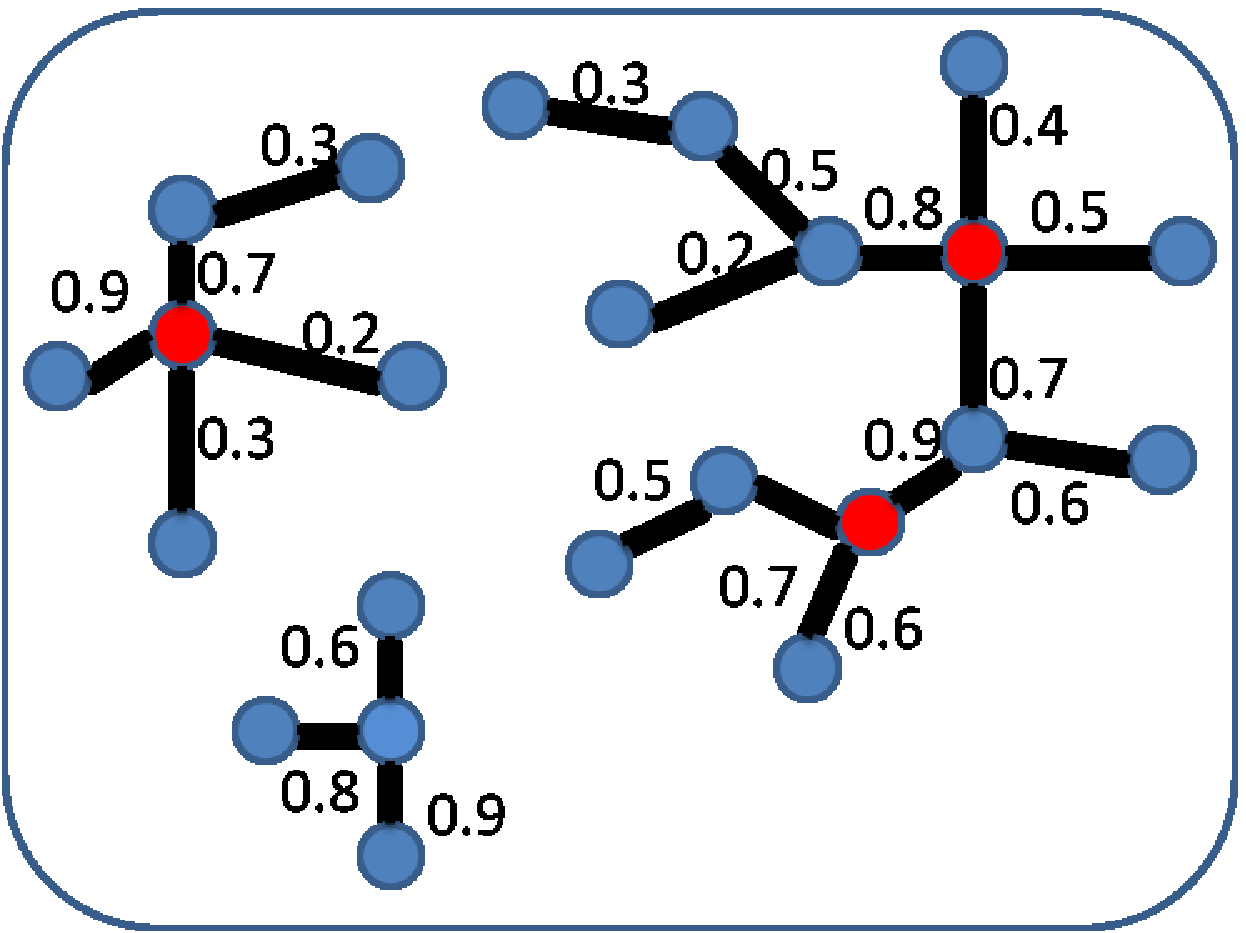}
}
\subfigure[D(A)=1.13]
{
\label{fig:SD:b}
\includegraphics[width=0.47\linewidth]{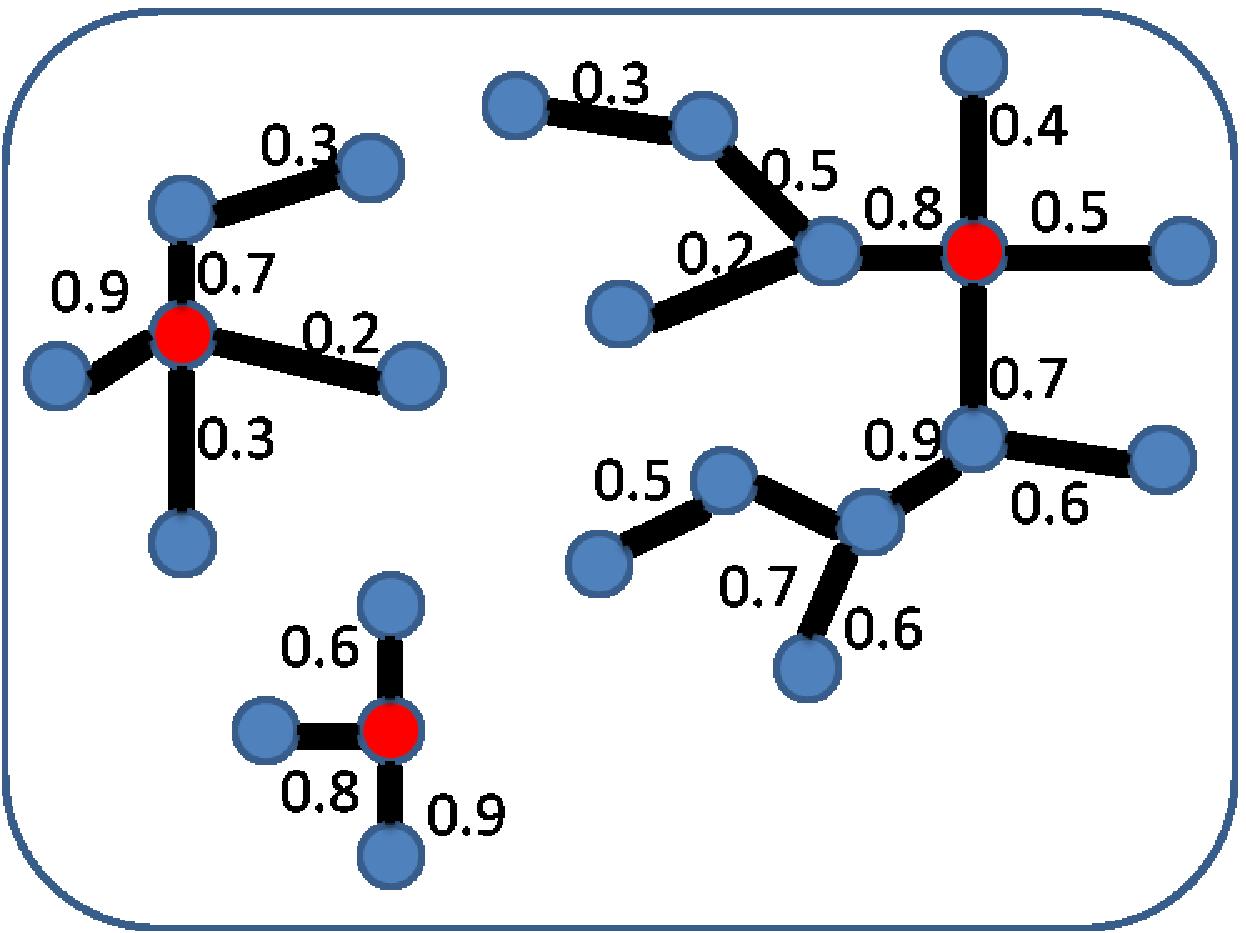}
}
\end{center}
%\vspace{-0.1cm}
\caption{The single-layer diversity term for the diverse proposal selection. Each node denotes a segment vertex (best viewed in color). Similarity between vertices are labelled next to each edge. The red node labels the selected segments. Each figure shows three exemplar clusters as connected groups. We can see the three exemplar clusters are unbalanced. Purely computing the weighted coverage term will pick the third node from the largest cluster to gain more similarity between the selected set and the whole set as in (a). While by computing the single-layer diversity term, we observe that (b) is preferred to (a) as it encourage diversity among the selected nodes.}
\label{fig:SD}
%\vspace{-0.1cm}
\end{figure}

\begin{eqnarray}
\label{eqn:diversity}
D(A)=\sum_{l=1}^{L}D_l(A)=\sum_{t,l}\sqrt{\sum_{j\in P_t^l \cap A} \frac{1}{|V^l|}(\sum_{i \in V^l} w_{ij})}
%D(A)&=&\sum_{l=1}^{L}D_l(A) \\
%&=&\sum_{t,l}\sqrt{\sum_{j\in P_t^l \cap A} \frac{1}{|V^l|}(\sum_{i \in V^l} w_{ij})} \nonumber
\end{eqnarray}
where $P_t^l$ is the set of segments which belong to cluster $t$ in layer $l$ (defined in section~\ref{sec:pce}). $|V^l|$ is the number of segments in layer $l$. This single-scale diversity term encourages $A$ to include elements from different clusters and leads to more diverse segments from each layer. The single-layer diversity term is submodular; a detailed proof is provided in the supplementary material. %by using the composition rule from Theorem 1. The square root is non-decreasing concave function. Within each square root is a non-negative weights. This non-negative weights is monotone non-decreasing as we add components into it. The square root of such non-negative weights yields a submodular function and summing them up retains submodularity. $D(A)$ is the summation of a series of submodular functions, thus it is submodular.
%As shown in Figure~\ref{fig:diversity}, the layer contains segments from background and foreground objects. There is a large amount of segments which has similar features as they all come from the same background. As single-scale diversity term encourages segments from different cluster, segment $v_3$ will also be chosen as it comes from a different cluster and contributes to diversity of selected set.

In many images, the background composes a large part of the image. For a single layer, the segments corresponding to objects are only a small percentage of all segments. The segment distributions corresponding to different objects and the background are generally unbalanced. The weighted coverage term favours selecting segments that well represent all segments, resulting in redundancy and occasionally missing small objects. Together with the single-layer diversity term, diversity among the selected segments are enforced as shown in Figure~\ref{fig:SD}.

\begin{figure}[t]
\begin{center}
   \includegraphics[width=0.99\linewidth]{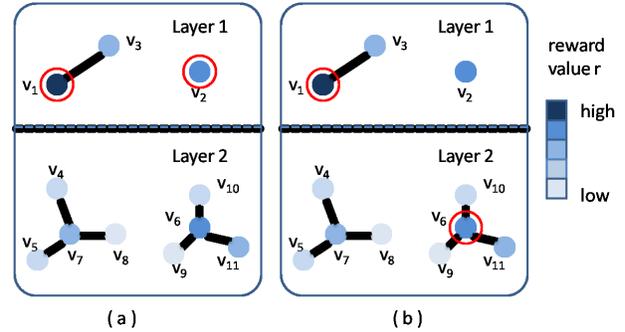}
\end{center}
\vspace{-0.2cm}
\caption{The multi-scale reward term for selecting proposals from different scales (best viewed in color). The nodes represent segments. The reward value $r_i$ of segment $v_i$ is reflected by color. The higher $r_i$, the more likely it is an object. The red circle denotes the selected nodes. Suppose $v_1$ has already been selected. We observe that $R\{v_1,v_2\}-R\{v_1\} <  R\{v_1,v_6\}-R\{v_1\}$. In another word, although $v_2$ and $v_6$ have similar reward value, $v_6$ from layer $2$ will brings higher marginal gain; thus $v_6$ is favoured over $v_2$ and (b) is preferred to (a).}
\label{fig:reward}
%\vspace{-0.1cm}
\end{figure}

\subsubsection{Multi-Scale Reward Term}
\label{subsubsec:mr}
Considering the multi-scale nature of objects in an image, we propose the following discriminative multi-scale reward term to encourage selected segments to have high likelihood of high object coverage. The multi-scale reward term is defined as:
\begin{eqnarray}
%\vspace{-0.4cm}
\label{eqn:reward}
R(A)=\sum_{l=1}^L\sqrt{\sum_{j\in V^l \bigcap A}r_j}
\end{eqnarray}
$V^l$ is the set of segments from layer $l$. The value $r_j$ estimates the likelihood of a segment to be an object. It determines the priority of a segment being chosen in its layer. We use CNN features to train a SVM model over object segments and non-object segments in training images and then assign a confidence score for each segment during testing. The confidence score is used as $r_j$ for a segment $v_j$. %Scores of each layer are normalized into the same scale. %One can also learn a different weight for different layer. Here we simply use equal weights for all layers.

%The multi-scale reward term encourages to select a set of discriminative segments from multi-scale segments generated from hierarchical segmentation. As soon as an element is selected from a layer, other elements from the same layer start to have diminishing gain because of the square root function. A simple example is shown in Figure~\ref{fig:reward}. For instance, consider the case where $v_1,v_2,v_4\in V^1$, $v_3, v_5, ... \in V^2$, and $r_{v_1}=0.70, r_{v_2}=0.60$, and $r_{v_3}=0.50$. Assume $v_1$ is already in the selected set $A$  (it is more representative based on salient segment select term). Greedily selecting the next element will choose $v_3$ rather than $v_2$ since $\sqrt{0.70+0.60}=\sqrt{1.30}=1.14 \leq \sqrt{0.70}+\sqrt{0.50}=1.54$. Similarly with $D(A)$, $R(A)$ can be proved to be submodular through the composition rule from Theorem 1. %As the square root is non-decreasing concave function. Within each square root is a non-negative weights. This non-negative weights is monotone non-decreasing as we add components into it. The square root of such non-negative weights yields a submodular function and summing them up retains submodularity.

The multi-scale reward term encourages $A$ to select a set of discriminative segments from multi-scale segments generated from a hierarchical segmentation. As soon as an element is selected from a layer, other elements from the same layer start to have diminishing gain because of the submodular property of $R(A)$. A simple example is shown in Figure~\ref{fig:reward}. Similar to $D(A)$, $R(A)$ is submodular and the proof is presented in the supplementary material.

%Intuitively, it will benefit more to select object proposals from a scale that has none of its elements already been chosen. For example, if we have two layers, each have more than 500 segments. If we already selected top 100 segments from layer 1, adding top 50 segments from layer 2 will bring more benefit than adding another 50 segments from layer 1.

%-------------------------------------------------------------------------
\section{Optimization}
We combine the weighted coverage term, the single-scale diversity term and the multi-scale reward term to find high-quality object proposals. The final objective function of object proposal generation is formulated as below:
\begin{eqnarray}
\label{eqn:obj}
\max_A F(A)&=&\max_{A}H(A)+\alpha D(A) + \beta R(A)\\
&=& \max_{A}\sum_{i\in V}\max_{j \in A}w_{ij}+ \beta \sum_{l=1}^L\sqrt{\sum_{j\in V^l \bigcap A}r_j} \nonumber \\
&& + \alpha \sum_{n,l}\sqrt{\sum_{j\in P_t^l \cap A} \frac{1}{|V^l|}(\sum_{i \in V^l} w_{ij})} \nonumber \\
&&s.t. \quad A \subseteq V, N_A\leq K, \alpha \geq 0, \beta \geq 0 \nonumber
%&&s.t. \quad A \subseteq J \subseteq V, N_A\leq K, \alpha, \beta \geq 0 \nonumber
\end{eqnarray}

The submodularity is preserved by taking non-negative linear combinations of the three submodular terms $H(A)$, $D(A)$, and $R(A)$. Direct maximization of equation~(\ref{eqn:obj}) is an NP-hard problem. We can approximately solve the problem via a greedy algorithm~\cite{Galvao04,Nemhauser78} based on its submodularity property. A lower bound of $(e-1)/e$ times the optimal value is guaranteed as proved in~\cite{Nemhauser78} (e is the base of the natural logarithm).

\begin{table}[b]
\begin{center}
\begin{tabular}{|l|l|l|l|l}
\cline{1-4}
& AUC  & Recall & BSS  &  \\ \cline{1-4}
C,T+layout~\cite{Endres14} & 77.5 & 83.4   & 67.2 &  \\ \cline{1-4}
all feature~\cite{Endres14} & 80.2 & 79.7   & 66.2 &  \\ \cline{1-4}
Ours & 81.1 & 83.6   & 71.8 &  \\ \cline{1-4}
\end{tabular}
\end{center}
\caption{Comparison of object proposals' quality on the BSDS dataset, measured with AUC, recall and BSS.}
\label{tab:bsds}
\end{table}

The algorithm starts from an empty set $A=\varnothing$. It adds the element $a^*$ which provides the largest marginal gain %$\rho$ for $F(A)$ 
among the unselected elements to $A$ iteratively. The iterations stop when $|A|$ reaches the desired capacity number $K$. % or when $F(A)$ starts decreasing. 
The optimization steps can be further accelerated using a lazy greedy approach from~\cite{Leskovec07}. Instead of recomputing gain for every unselected element after each iteration, an ordered list of marginal benefits will be maintained in descending order. Only the top unselected segment is re-evaluated at each iteration. Other unselected segments will be re-evaluated only if the top segment does not remain at the top after re-evaluation. The pseudo code is presented in Algorithm \ref{algorithm:alg1}.

%\input{submodularAlgorithm}
%\begin{algorithm}
%  \caption{Submodular object proposal generation}
%  \label{algorithm:alg1}
 % \begin{algorithmic}
%  	\State \textbf{Input:} $I$, $G=(V,E)$, $K$, $\alpha$, $\beta$
 % 	\State \textbf{Output:} $A$
 % 	\State Initialization: $A=\varnothing$
 %   \While  {$|A|<K$}
 %   \State $a^*=arg\max_{\{A\cup a\}} F(A\cup \{a\})-F(A)$
%    \If{$F(A\cup \{a^*\})-F(A)\leq F(A)$}
%    \State break
%    \EndIf
%    \State $A \leftarrow A\cup {a^*}$ %, $\rho_{a^*}=0$
%    \For {$i \in  V \setminus A$}
%    	\State $\rho_i=\rho_i^{new}$
%    	\EndFor
%    \EndWhile
 %   \State load corresponding segment maps based on index set $A$
%  \end{algorithmic}
%\end{algorithm}
\begin{algorithm}
  \caption{Submodular object proposal generation}
  \label{algorithm:alg1}
  \begin{algorithmic}
  	\State \textbf{Input:} $I$, $G=(V,E)$, $K$, $\alpha$, $\beta$
  	\State \textbf{Output:} $A$
  	\State Initialization: $A \leftarrow \varnothing$, $U \leftarrow V$
    \Loop 
    \State $a^*=arg \underset{a\in U}{\max} F(A\cup \{a\})-F(A)$
    \If{$|A| \geq K$}
    \State break
    \EndIf
    \State $A \leftarrow A\cup \{a^*\}$ 
    \State $U \leftarrow U - \{a^*\}$ 
%    \For {$i \in  V \setminus A$}
%    	\State $\rho_i=\rho_i^{new}$
%    	\EndFor
    \EndLoop
 %   \State load corresponding segment maps based on index set $A$
  \end{algorithmic}
\end{algorithm}
%\vspace{-0.2cm}

%------------------------------------------------------------------------
%\input{experiment}
\section{Experiments}
We evaluate our approach on two public datasets: BSDS~\cite{Martin02} and PASCAL VOC2012~\cite{pascal-voc-2012} segmentation dataset. The results for PASCAL VOC2012 are on the validation set of the segmentation task. We evaluate the object proposal quality by assessing the best proposal for each object using the Jaccard index score (see details in section~\ref{subsec:objEva}). We also compare our ranking method with several baselines~\cite{Endres14} and analyses the efficiency of our object proposals on the object recognition task. 
%\vspace{-0.6cm}

\begin{table*}\footnotesize
% \tiny
\begin{center}
% \begin{tabular}{|L{0.35cm}|C{0.15cm}|C{0.3cm} C{0.3cm} C{0.3cm} C{0.3cm} C{0.3cm} C{0.3cm} C{0.3cm} C{0.3cm} C{0.3cm} C{0.3cm} C{0.3cm} C{0.3cm} C{0.3cm} C{0.3cm} C{0.3cm} C{0.3cm} C{0.3cm} C{0.3cm} C{0.3cm} C{0.3cm} |C{0.2cm}|}
% \hline
% Method & N & Plane & Bike & Bird & Boat & Bottle & Bus & Car & Cat & Chair & Cow & Table & Dog & Horse & MBike & Person & Plant & Sheep & Sofa & Train & TV & Global \\
\begin{tabular}{cc|p{0.22cm}p{0.22cm}p{0.22cm}p{0.22cm}p{0.22cm}p{0.22cm}p{0.22cm}p{0.22cm}p{0.22cm}p{0.22cm}p{0.22cm}p{0.22cm}p{0.22cm}p{0.22cm}p{0.22cm}p{0.22cm}p{0.22cm}p{0.22cm}p{0.22cm}p{0.22cm}p{0.22cm}p{0.22cm}c}
\hline
{Method}&{N}&{\begin{sideways}Plane\end{sideways}}&{\begin{sideways}Bike\end{sideways}}&{\begin{sideways}Bird\end{sideways}}&{\begin{sideways}Boat\end{sideways}}&{\begin{sideways}Bottle\end{sideways}}&{\begin{sideways}Bus\end{sideways}}&{\begin{sideways}Car\end{sideways}}&{\begin{sideways}Cat\end{sideways}}&{\begin{sideways}Chair\end{sideways}}&{\begin{sideways}Cow\end{sideways}}&{\begin{sideways}Table\end{sideways}}&{\begin{sideways}Dog\end{sideways}}&{\begin{sideways}Horse\end{sideways}}&{\begin{sideways}MBike\end{sideways}}&{\begin{sideways}Person\end{sideways}}&{\begin{sideways}Plant\end{sideways}}&{\begin{sideways}Sheep\end{sideways}}&{\begin{sideways}Sofa\end{sideways}}&{\begin{sideways}Train\end{sideways}}&{\begin{sideways}TV\end{sideways}}&{\begin{sideways}Global\end{sideways}} \\
\hline
Ours & 1100 & \textbf{82.3} & 48.8 & \textbf{84.6} & \textbf{76.7} & \textbf{71.4} & \textbf{80.6} & 67.7 & \textbf{93.1} & \textbf{69.7} & \textbf{86.0} & 78.5 & \textbf{89.7} & \textbf{83.2} & 77.3 & 72.9 & \textbf{70.4} & 77.8 & \textbf{85.8} & \textbf{85.0} & \textbf{87.5} & \textbf{76.5} \\
\cite{Arbelaez14} & 1100 & 80.0 & 47.8 & 83.9 & 76.4 & 71.1 & 78.5 & \textbf{68.9} & 89.3 & 68.5 & 85.9 & 79.8 & 85.8 & 80.4 & 75.4 & \textbf{73.5} & 69.3 & \textbf{84.9} & 82.6 & 81.7 & 85.8 & 76.0 \\
\cite{Endres14} & 1100 & 75.1 & \textbf{49.1} & 80.7 & 68.8 & 62.8 & 76.4 & 63.3 & 89.4 & 64.6 & 83.0 & \textbf{80.3} & 83.7 & 78.4 & \textbf{78.0} & 66.9 & 66.2 & 69.5 & 82.0 & 84.3 & 81.8 & 71.6 \\
\cite{Arbelaez12} & 1100 & 74.4 & 46.6 & 80.5 & 69.4 & 64.6 & 73.5 & 61.2 & 89.0 & 65.1 & 80.5 & 78.4 & 85.2 & 77.2 & 70.6 & 67.9 & 68.8 & 73.5 & 81.6 & 75.8 & 82.0 & 71.4 \\
\cite{Kim12} & 1100 & 73.8 & 40.6 & 75.8 & 66.7 & 52.7 & 79.7 & 50.6 & 91.2 & 59.2 & 80.2 & 80.7 & 87.4 & 79.0 & 74.7 & 62.1 & 54.6 & 65.0 & 84.6 & 82.4 & 79.5 & 67.4 \\
\cite{Sande11} & 1100 & 68.3 & 39.6 & 70.6 & 64.8 & 58.0 & 68.2 & 51.8 & 77.6 & 58.2 & 72.6 & 70.4 & 74.0 & 66.2 & 59.9 & 59.8 & 55.4 & 67.7 & 71.3 & 68.6 & 78.7 & 63.1 \\
\hline
ours & 100 & \textbf{75.2} & \textbf{40.8} & \textbf{78.4} & \textbf{70.3} & \textbf{55.5} & \textbf{72.8} & \textbf{51.1} & \textbf{83.4} & \textbf{56.8} & 77.3 & 66.7 & \textbf{84.4} & \textbf{75.2} & 65.9 & 59.3 & \textbf{54.9} & 68.1 & \textbf{77.9} & \textbf{76.1} & 76.8 & \textbf{64.3} \\
\cite{Arbelaez14} & 100 & 70.2 & 38.8 & 73.6 & 67.7 & 55.3 & 68.5 & 50.6 & 82.4 & 54.4 & \textbf{78.1} & 67.7 & 77.7 & 69.3 & 66.3 & \textbf{59.9} & 51.4 & \textbf{70.2} & 74.1 & 72.6 & \textbf{78.1} & 63.7 \\
\cite{Endres14} & 100 & 70.6 & \textbf{40.8} & 74.8 & 59.9 & 49.6 & 65.4 & 50.4 & 81.5 & 54.5 & 74.9 & \textbf{68.1} & 77.3 & 69.3 & \textbf{66.8} & 56.2 & 54.3 & 64.1 & 72.0 & 71.6 & 69.9 & 61.7 \\
\cite{Carreira12} & 100 & 72.7 & 36.2 & 73.6 & 63.3 & 45.4 & 67.4 & 39.5 & 84.1 & 47.7 & 73.2 & 64.0 & 81.1 & 72.2 & 64.3 & 52.8 & 42.9 & 62.2 & 72.9 & 74.3 & 69.5 & 59.0 \\
\hline
\end{tabular}
\end{center}
\caption{VOC2012 val set. Jaccard index at the instance level and class level.}
\label{tab:bssPascal}
\end{table*}

\begin{figure}
%\vspace{-0.6cm}
\begin{center}
   \includegraphics[width=0.9\linewidth]{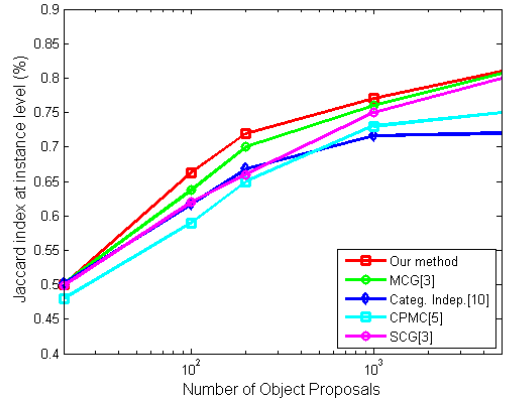}
\end{center}
%\vspace{-0.1cm}
\caption{Object proposal quality on PASCAL VOC2012 validation set, measured with the Jaccard index at instance level $J_i$.}
\label{fig:pascalBSS}
%\vspace{-0.1cm}
\end{figure}

\begin{figure*}
\begin{center}
   \includegraphics[width=0.99\linewidth]{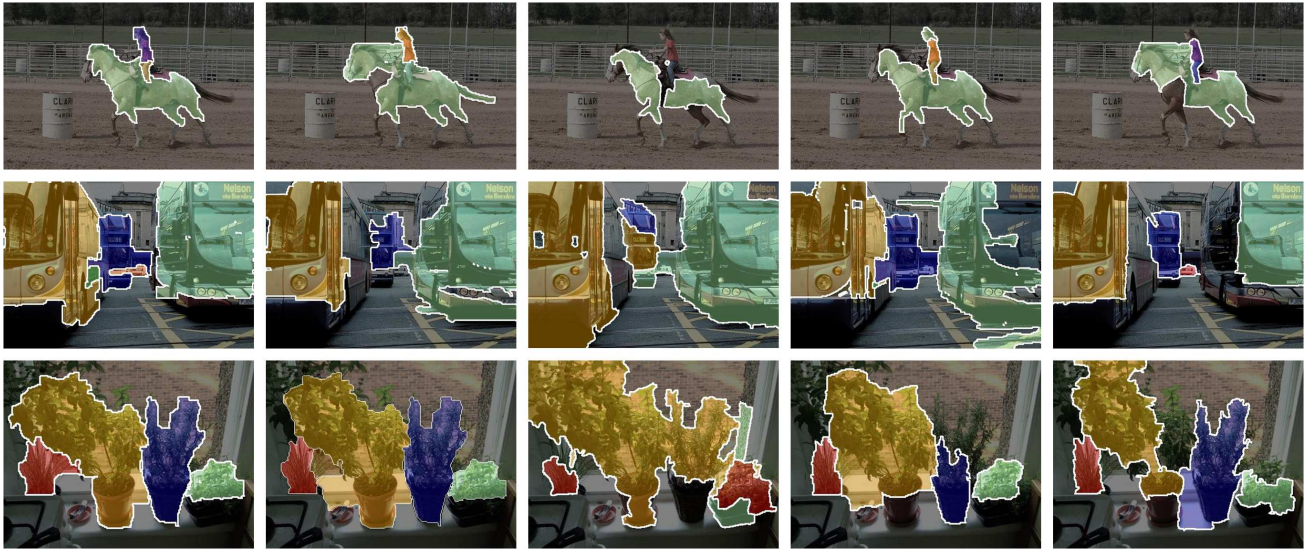}
\end{center}
%\vspace{-0.1cm}
\caption{Sample object proposals from the PASCAL VOC2012. The left column shows the best four proposals for objects in our model. The remaining columns show the highest ranked proposals with at least 50 percent overlap with an object. The second column is from our method, the third column is from Categ. Indep.~\cite{Endres14}, the fourth column is from CPMC~\cite{Carreira12}, and the last column is from MCG~\cite{Arbelaez14}.}
\label{fig:obj}
%\vspace{-0.1cm}
\end{figure*}

\subsection{Proposal evaluation}
\label{subsec:objEva}
To measure the quality of a set of object proposals, we followed~\cite{Arbelaez14} and compute the Jaccard index score, or the best segmentation overlap score (BSS) for each object. The overall quality of a object proposal set is measured at the class level and the instance level. The Jaccard index at instance level, denoted as $J_i$, is defined as the mean of BSS over all objects. The Jaccard index at class level, $J_c$ is defined as the mean of BSS over objects from each category. 

\subsubsection{BSDS dataset}
We compare our object proposals with~\cite{Endres14}. For fair comparison, we also compute the area under the ROC curve (AUC) and recall defined with an overlap threshold at 50 per cent. The results are summarized in Table~\ref{tab:bsds}. Our object proposal achieves the best performance.

\subsubsection{PASCAL VOC2012}
\begin{table*}\footnotesize
\begin{center}
\begin{tabular}{c|p{0.22cm}p{0.22cm}p{0.22cm}p{0.22cm}p{0.22cm}p{0.22cm}p{0.22cm}p{0.22cm}p{0.22cm}p{0.22cm}p{0.22cm}p{0.22cm}p{0.22cm}p{0.22cm}p{0.22cm}p{0.22cm}p{0.22cm}p{0.22cm}p{0.22cm}p{0.22cm}p{0.22cm}p{0.22cm}c}
\hline
{Method}&{\begin{sideways}Plane\end{sideways}}&{\begin{sideways}Bike\end{sideways}}&{\begin{sideways}Bird\end{sideways}}&{\begin{sideways}Boat\end{sideways}}&{\begin{sideways}Bottle\end{sideways}}&{\begin{sideways}Bus\end{sideways}}&{\begin{sideways}Car\end{sideways}}&{\begin{sideways}Cat\end{sideways}}&{\begin{sideways}Chair\end{sideways}}&{\begin{sideways}Cow\end{sideways}}&{\begin{sideways}Table\end{sideways}}&{\begin{sideways}Dog\end{sideways}}&{\begin{sideways}Horse\end{sideways}}&{\begin{sideways}MBike\end{sideways}}&{\begin{sideways}Person\end{sideways}}&{\begin{sideways}Plant\end{sideways}}&{\begin{sideways}Sheep\end{sideways}}&{\begin{sideways}Sofa\end{sideways}}&{\begin{sideways}Train\end{sideways}}&{\begin{sideways}TV\end{sideways}}&{\begin{sideways}Mean\end{sideways}} \\
\hline
O$_2$P~\cite{Hariharan14} & 56.5 & 19.0 & 23.0 & 12.2 & 11.0 & 48.8 & 26.0 & 43.3 & 4.7 & 15.6 & 7.8 & 24.2 & 27.5 & 32.3 & 23.5 & 4.6 & 32.3 & 20.7 & 38.8 & 32.3 & 25.2 \\
\hline
SDS-A~\cite{Hariharan14} & 61.8 & 43.4 & 46.6 & 27.2 & 28.9 & 61.7 & 46.9 & 58.4 & 17.8 & 38.8 & 18.6 & 52.6 & 44.3 & 50.2 & 48.2 & 23.8 & 54.2 & 26.0 & 53.2 & 55.3 & 42.9 \\
\hline
SDS-B~\cite{Hariharan14} & 65.7 & \textbf{49.6} & 47.2 & 30.0 & 31.7 & \textbf{66.9} & 50.9 & 69.2 & 19.6 & 42.7 & 22.8 & 56.2 & 51.9 & 52.6 & 52.6 & 25.7 & \textbf{54.2} & \textbf{32.2} & 59.2 & 58.7 & 47.0 \\
\hline
SDS-C~\cite{Hariharan14} & 67.4 & \textbf{49.6} & 49.1 & 29.9 & 32.0 & 65.9 & \textbf{51.4} & \textbf{70.6} & \textbf{20.2} & 42.7 & 22.9 & 58.7 & 54.4 & 53.5 & 54.4 & 24.9 & 54.1 & 31.4 & 62.2 & 59.3 & 47.7 \\
\hline
Ours & \textbf{68.2} & 14.0 & \textbf{64.7} & \textbf{51.3} & \textbf{39.3} & 62.1 & 45.6 & 65.8 & 9.9 & \textbf{49.1} & \textbf{30.8} & \textbf{61.9} & \textbf{54.9} & \textbf{65.9} & \textbf{54.5} & \textbf{31.8} & 48.4 & 29.5 & \textbf{73.9} & \textbf{65.6} & \textbf{48.9} \\
\hline
\end{tabular}
\end{center}
\caption{Results on AP$^r$ on the PASCAL VOC2012 val. All numbers are $\%$.}
\label{tab:sdsseg}
%\vspace{-0.2cm}
\end{table*}

We evaluate our object proposal approach on the PASCAL VOC2012 validation dataset. The SVM classifier for reward value (details in section~\ref{subsubsec:mr}) is trained on the training dataset. Our object proposals are compared with~\cite{Kim12, Carreira12, Arbelaez12, Sande11, Endres14, Arbelaez14}. As shown in Table~\ref{tab:bssPascal}, our method outperform all other methods with the same number of object proposals for Jaccard index at the instance level. Meanwhile, we achieve the highest scores on most of the classes (14 out of 20). In Figure~\ref{fig:pascalBSS}, we show how $J_i$ changes as the number of object proposals increases. Since our approach prefers to select representative, diverse and multi-scale object proposals, our proposal quality outperform MCG~\cite{Arbelaez14}, Categ. Indep.~\cite{Endres14}, CPMC~\cite{Carreira12}, and SCG~\cite{Arbelaez14} with only a small number of proposals. In Figure~\ref{fig:obj}, we show some qualitative results of our object proposals. We observe that our proposals can capture diverse objects of different sizes. In addition, we compare our proposal generation time with MCG~\cite{Arbelaez14} which also uses multi-scale information. Our method takes about 7 seconds per image compared to 10 seconds reported in~\cite{Arbelaez14}. The parameters are set $\alpha=3.9$, $\beta=2.0$ in our experiments.

\begin{table*}\footnotesize
\begin{center}
\begin{tabular}{c|p{0.22cm}p{0.22cm}p{0.22cm}p{0.22cm}p{0.22cm}p{0.22cm}p{0.22cm}p{0.22cm}p{0.22cm}p{0.22cm}p{0.22cm}p{0.22cm}p{0.22cm}p{0.22cm}p{0.22cm}p{0.22cm}p{0.22cm}p{0.22cm}p{0.22cm}p{0.22cm}p{0.22cm}p{0.22cm}c}
\hline
{Method}&{\begin{sideways}Plane\end{sideways}}&{\begin{sideways}Bike\end{sideways}}&{\begin{sideways}Bird\end{sideways}}&{\begin{sideways}Boat\end{sideways}}&{\begin{sideways}Bottle\end{sideways}}&{\begin{sideways}Bus\end{sideways}}&{\begin{sideways}Car\end{sideways}}&{\begin{sideways}Cat\end{sideways}}&{\begin{sideways}Chair\end{sideways}}&{\begin{sideways}Cow\end{sideways}}&{\begin{sideways}Table\end{sideways}}&{\begin{sideways}Dog\end{sideways}}&{\begin{sideways}Horse\end{sideways}}&{\begin{sideways}MBike\end{sideways}}&{\begin{sideways}Person\end{sideways}}&{\begin{sideways}Plant\end{sideways}}&{\begin{sideways}Sheep\end{sideways}}&{\begin{sideways}Sofa\end{sideways}}&{\begin{sideways}Train\end{sideways}}&{\begin{sideways}TV\end{sideways}}&{\begin{sideways}Mean\end{sideways}} \\
% & Plane & Bike & Bird & Boat & Bottle & Bus & Car & Cat & Chair & Cow & Table & Dog & Horse & MBike & Person & Plant & Sheep & Sofa & Train & TV & Mean \\
\hline
O$ _2$P~\cite{Hariharan14} & 46.8 & 21.2 & 22.1 & 13.0 & 10.1 & 41.9 & 24.0 & 39.2 & 6.7 & 14.6 & 9.9 & 24.0 & 24.4 & 28.6 & 25.6 & 7.0 & 29.0 & 18.8 & 34.6 & 25.9 & 23.4 \\
\hline
SDS-A~\cite{Hariharan14} & 48.3 & 39.8 & 39.2 & 25.1 & 26.0 & 49.5 & 39.5 & 50.7 & 17.6 & 32.5 & 18.5 & 46.8 & 37.7 & 41.1 & 43.2 & 23.4 & 43.0 & 26.2 & 45.1 & 47.7 & 37.0 \\
\hline
SDS-B~\cite{Hariharan14} & 51.1 & 42.1 & 40.8 & 27.5 & 26.8 & 53.4 & 42.6 & 56.3 & 18.5 & 36.0 & 20.6 & 48.9 & 41.9 & 43.2 & 45.8 & 24.8 & 44.2 & 29.7 & 48.9 & 48.8 & 39.6 \\
\hline
SDS-C~\cite{Hariharan14} & 53.2 & 42.1 & 42.1 & 27.1 & 27.6 & 53.3 & 42.7 & 57.3 & 19.3 & 36.3 & 21.4 & 49.0 & 43.6 & 43.5 & 47.0 & 24.4 & 44.0 & 29.9 & 49.9 & 49.4 & 40.2 \\
\hline
SDS-C+ref~\cite{Hariharan14} & 52.3 & \textbf{42.6} & 42.2 & 28.6 & 28.6 & \textbf{58.0} & \textbf{45.4} & 58.9 & \textbf{19.7} & 37.1 & 22.8 & 49.5 & 42.9 & 45.9 & \textbf{48.5} & 25.5 & \textbf{44.5} & \textbf{30.2} & 52.6 & 51.4 & 41.4 \\
\hline
Ours & \textbf{54.7} & 19.4 & \textbf{54.3} & \textbf{40.9} & \textbf{34.4} & 52.0 & 41.3 & \textbf{59.3} & 13.3 & \textbf{42.9} & \textbf{25.8} & \textbf{51.9} & \textbf{44.8} & \textbf{51.5} & 47.0 & \textbf{31.4} & 42.6 & 28.5 & \textbf{59.2} & \textbf{53.8} & \textbf{42.4} \\
\hline
\end{tabular}
\end{center}
\caption{Results on AP$^r_{vol}$ on the PASCAL VOC2012 val. All numbers are $\%$.}
\label{tab:sdsaprvol}
\end{table*}

\begin{figure}
%\vspace{-0.2cm}
\begin{center}
   \includegraphics[width=0.95\linewidth]{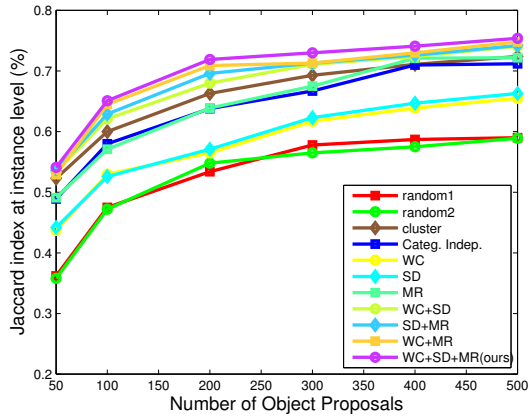}
\end{center}
%\vspace{-0.1cm}
\caption{Comparing different ranking methods (random selection, clustering, Categ. Indep.~\cite{Endres14}, WC, SD, MR, WC+SD, WC+MR, SD+MR, WC+SD+MR(ours)).}
\label{fig:rank}
%\vspace{-0.1cm}
\end{figure}

\begin{figure*}
\begin{center}
   \includegraphics[width=0.85\linewidth]{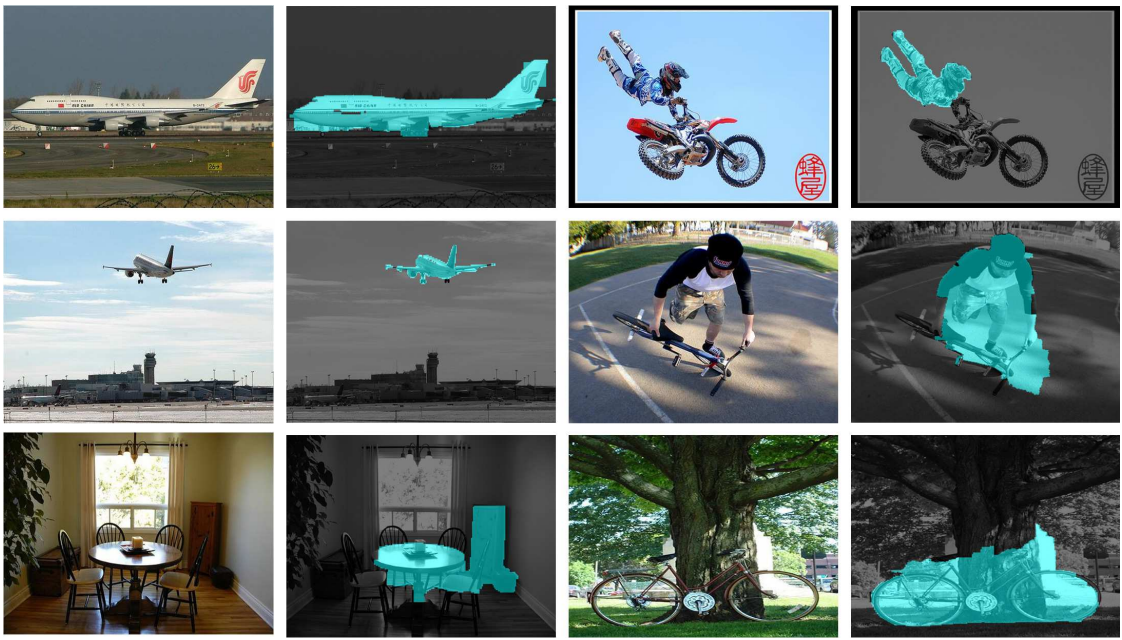}
\end{center}
%\vspace{-0.1cm}
\caption{Top detections on: aeroplane, person, dining table, bicycle.Our detection results work well on objects of different scales.}
\label{fig:sds}
%\vspace{-0.1cm}
\end{figure*}

\subsection{Ranking performance}
\label{subsec:rank}
To explore our method's ranking ability, we compare our ranking method with four baselines on the PASCAL VOC2012 dataset. 1) \textbf{Random1} randomly selects object proposals from the multi-scale segment pool. 2) \textbf{Random2} randomly selects object proposals from each layer evenly, and combine them together. 3) \textbf{Clustering} selects the object proposals which are closest to the cluster center based on euclidean distance. The cluster centres are obtained via k-means clustering and k is set to be the number of object proposals to be selected. 4) \textbf{Categ. Indep.} is the method from~\cite{Endres14} to rank segments. In order to show the importance of each term in our model, we evaluate each term: the weighted coverage term(WC), the single-layer diversity term (SD), and the multi-scale reward term (MR). Results of different term combinations (WC+SD, WC+MR, SD+MR) and the full model (WC+SD+MR) are also presented.

Figure~\ref{fig:rank} shows the quality of the selected object proposals
% ($J_i$)
 using different ranking methods from the same segment pool. The two random selection methods achieve similar object proposal qualities. Comparing WC, SD and MR terms independently, WC achieves lower quality than the other two. As discussed in~\ref{subsubsec:wc}, it emphasize the representativeness of the selected set regardless of whether the segment is an object or not. The clustering method also has the same weakness. The MR term is comparable to structured learning as it also takes into account multi-scale information. Adding the MR term to each of the WC and SD terms increases performance as it introduces discriminative information into the proposal selection process. Our full ranking model selects the best object proposals amongst all. 

\subsection{Semantic Segmentation and Object Detection}
To analyse the utility of the object proposals generated by our approach in real object recognition tasks, we perform semantic segmentation and object detection on the PASCAL VOC2012 validation set. We follow the settings in~\cite{Hariharan14}, where 2000 object proposals are generated for each image using our algorithm. Then we extract CNN features for both the regions and their bounding boxes using the deep convolutional neural network model pre-trained on ImageNet and fine-tuned on the PASCAL VOC2012 training set, the same as in~\cite{Hariharan14}. These features are concatenated, then passed through linear classifiers trained for region and box classification tasks. After non-maxima suppression, we select the top 20,000 detections for each category.

The results are evaluated with the traditional bounding box AP$^b$ and the extended metric AP$^r$ as in~\cite{Hariharan14} (the superscripts $b$ and $r$ correspond to region and bounding box). 
%AP$^r$ and AP$^r_{vol}$ measures~\cite{Hariharan14}. 
The AP$^r$ score is the average precision of whether a hypothesis overlaps with the ground-truth instance by over $50\%$, and the AP$^r_{vol}$ is the volume under the precision recall (PR) curve, which are suitable for the simultaneous segmentation and detection task. The evaluation of the detection task uses $AP^b$ and $AP^b_{vol}$, which are conventional evaluation metric for object detection. %The evaluation of the detection task uses similar measures, denoted as the $AP^b$ and $AP^b_{vol}$, where $AP^b$ is the conventional evaluation metric for object detection.

Table~\ref{tab:sdsseg} and Table~\ref{tab:sdsaprvol} shows the AP$^r$ and AP$^r_{vol}$ results for each class. We can see that the results using our object proposals, both our mean AP$^r$ and mean AP$^r_{vol}$ have achieved state of the art using a seven-layer network, and we outperform previous methods in 14 out of 20 classes. In contrast to SDS~\cite{Hariharan14}, we neither fine tune different networks for regions and boxes nor refine the regions after classification. But our results still not only outperform the corresponding SDS-A but also the complicated SDS-B and SDS-C methods which finetuned two networks separately and as a whole. Moreover, on the more meaningful measurement of AP$^r_{vol}$ shown in Table~\ref{tab:sdsaprvol}, results based on our object proposals even outperform that of SDS-C+ref, where the segments are refined within their $10\times 10$ grid using a pretrained model with class priors. It shows the importance of good quality regions even before carefully designed feature extraction and region refinement after classification. 

Table~\ref{tab:sdsdet} shows the mean AP$^b$ and mean AP$^b_{vol}$ results for object detection. We achieved better results than RCNN~\cite{Girshick14CVPR}, RCNN-MCG~\cite{Hariharan14} and SDS-A~\cite{Hariharan14}, which shows that better region proposals not only improve segmentation but also give better localization of objects. Figure~\ref{fig:sds} shows some examples of our detection results. %Our detection results work well on objects of different scales. 
\begin{table}
\begin{center}
\begin{tabular}{|c|c|c|c|c|}
\hline
 & RCNN & RCNN-MCG & SDS-A & Ours \\
\hline
mean AP$^b$ & 51.0 & 51.7 & 51.9 & \textbf{52.4} \\
\hline
mean AP$^b_{vol}$ & 41.9 & 42.4 & 43.2 & \textbf{44.3} \\
\hline
\end{tabular}
\end{center}
\caption{Results on AP$^b$ and AP$^b_{vol}$ on the PASCAL VOC2012 val. All numbers are $\%$.}
\label{tab:sdsdet}
%\vspace{-0.4cm}
\end{table}

%------------------------------------------------------------------------
\section{Conclusion}
%\vspace{-0.6cm}
We presented an efficient approach to extract multi-scale object proposals. Built on the top of hierarchical image segmentation, exemplar clusters are first generated within each scale to discover different object patterns. By introducing a weighted coverage term, a single-scale diversity term and a multi-scale reward term, we define a submodular objective function to select object proposals from multiple scales. The problem is solved using a highly efficient greedy algorithm with guaranteed performance. The experimental results on the BSDS dataset and the PASCAL VOC2012 dataset demonstrate that our method achieves state-of-art performance and is computationally efficient. We further evaluate our object proposals on a simultaneous detection and segmentation task to demonstrate the effectiveness of our approach and outperform the object proposals generated by other methods.

{\small
\bibliographystyle{ieee}
\bibliography{egbib}
}

\end{document}